\documentclass{article}

\usepackage{microtype}
\usepackage{graphicx}
\usepackage{booktabs} 

\usepackage[accepted]{icml2023}

\icmltitlerunning{The Unintended Consequences of Discount Regularization}

\usepackage{graphicx}
\usepackage{amsthm}
\usepackage{apalike}
\usepackage{amssymb}
\usepackage{algorithm}
\usepackage{algorithmic}
\usepackage{outlines}
\usepackage{mathtools}
\usepackage{enumitem}
\setlist[]{nolistsep}
\usepackage{caption}
\usepackage{subcaption} 
\usepackage{soul}
\newtheorem{theorem}{Theorem}

\begin{document}

\twocolumn[
\icmltitle{The Unintended Consequences of Discount Regularization: \\Improving Regularization in Certainty Equivalence Reinforcement Learning}



\icmlsetsymbol{equal}{*}

\begin{icmlauthorlist}
\icmlauthor{Sarah Rathnam}{h}
\icmlauthor{Sonali Parbhoo}{i}
\icmlauthor{Weiwei Pan}{h}
\icmlauthor{Susan A. Murphy}{h}
\icmlauthor{Finale Doshi-Velez}{h}
\end{icmlauthorlist}

\icmlaffiliation{h}{Harvard University, School of Engineering and Applied Sciences, Cambridge, MA USA}
\icmlaffiliation{i}{Imperial College London, London UK}
\icmlcorrespondingauthor{Sarah Rathnam}{sarah\_rathnam@g.harvard.edu}

\icmlkeywords{Machine Learning, ICML, Reinforcement Learning, Regularization}

\vskip 0.3in
]



\printAffiliationsAndNotice{}  

\begin{abstract}
Discount regularization, using a shorter planning horizon when calculating the optimal policy, is a popular choice to restrict planning to a less complex set of policies when estimating an MDP from sparse or noisy data \citep{jiang2015dependence}.  It is commonly understood that discount regularization functions by de-emphasizing or ignoring delayed effects.  In this paper, we reveal an alternate view of discount regularization that exposes unintended consequences.  We demonstrate that planning under a lower discount factor produces an identical optimal policy to planning using any prior on the transition matrix that has the same distribution for all states and actions.  In fact, it functions like a prior with stronger regularization on state-action pairs with more transition data. This leads to poor performance when the transition matrix is estimated from data sets with uneven amounts of data across state-action pairs. Our equivalence theorem leads to an explicit formula to set regularization parameters locally for individual state-action pairs rather than globally. We demonstrate the failures of discount regularization and how we remedy them using our state-action-specific method across simple empirical examples as well as a medical cancer simulator.

\end{abstract}

\section{Introduction}

In reinforcement learning (RL), planning under a shorter horizon is a common form of regularization. In the most extreme case, a discount factor of zero results in a contextual bandit setting. Using a reduced or zero discount factor for planning is common in real-world applications such as mobile health (\citet{liao2020personalized}, \citet{trella2022reward}), medicine (\citet{oh2022reinforcement}, \citet{awasthi2022vacsim}, \citet{durand2018contextual}), and education (\citet{cai2021bandit}, \citet{qi2018bandit}).

In this paper, we analyze discount regularization in the context of \emph{certainty equivalence RL}.  This means that the agent takes the estimated model as true when calculating the optimal policy \cite{goodwin1984adaptive}. While planning using a reduced discount factor leads to better-performing policies in many cases \cite{jiang2015dependence, amit2020discount}, our main contribution is to present a deeper conception of this method that reveals limitations. We do so by first proving that discount regularization produces the same optimal policy as averaging the transition matrix for each action with a transition matrix in which all rows are the same. This can also be viewed in terms of a prior on the transition matrix. 

As further contributions, we utilize our reframing to expose unintended consequences. One such consequence is that the magnitude of the prior implied by discount regularization is higher for state-action pairs with more transition observations in the data and vice versa.  This is generally not desirable as we want stronger regularization on states that we have observed less, and want to rely on the data in states where we have more. 
Another negative aspect we expose is the assumption of equal transition distributions for all state-action pairs, which is inappropriate in many contexts. 

We also offer solutions to the problems exposed above in order to tailor regularization to the task at hand-- both the data set and the environment. To mitigate the issue of inconsistent prior magnitudes in data sets with uneven exploration, we derive a state-action specific formula for the regularization parameter. Furthermore, the method by which we derive this parameter can be adapted to other priors to match the transition dynamics of the environment. 

Finally, we demonstrate our results empirically on tabular examples and on a medical cancer dynamics simulator. First, we empirically confirm that discount regularization and a uniform prior on the transition matrix yield identical optimal policies. We then demonstrate that a uniform prior with fixed magnitude across state-action pairs outperforms discount regularization across environments. We also show that our state-action-specific regularization parameter reduces loss without parameter tuning.


\section{Related Works}\label{sec:relatedworks}

\citet{jiang2015dependence} demonstrate that planning under a shorter horizon often yields policies that outperform ones learned using the true discount factor, even when both are evaluated in the true environment. They prove that using a lower planning discount factor restricts the planning to a less complex set of policies, thereby avoiding overfitting. They further demonstrate that the benefit of a lower discount factor is increasingly pronounced in cases where the model is estimated from a smaller data set.  \citet{amit2020discount} refer to this concept as “discount regularization,” a term which we use here. Unlike these works, we provide means to connect discount regularization with placing a prior on the transition matrix. 

While a Bayesian prior encodes expert knowledge, information from previous studies, or other outside information, we can also view a prior as a form of regularization since it forces the model not to overfit when data is limited \citep{poggio1990networks,ghavamzadeh2015bayesian}. This is a flexible tool that allows us to regularize in a way that matches our prior knowledge and beliefs about the environment. 
In model-based Bayesian RL, the problem is often framed as a Bayes-Adaptive MDP (BAMDP), an MDP where the states are replaced by ``hyperstates'' that reflect the original state space combined with the posterior parameters of the transition function \cite{duff2002optimal}. In general, Bayesian RL algorithms do not explicitly address planner overfitting; rather they incorporate the probability distribution over models, causing the planner not to overfit to an uncertain model. For example, model-based Bayesian RL methods draw sample models from the posterior \citep{asmuth2012bayesian}, sample hyperstates \citep{poupart2006analytic}, or apply an exploration bonus based on the amount of data \citep{kolter2009near} or based on the variance of the parameters \citep{sorg2012variance}. The BAMDP framework can also be extended to the case of partial observability \cite{ross2007bayes,ross2011bayesian}. In this paper, we consider planning using the posterior mean of the transition matrix under a Dirichlet prior as a regularized form of the transition matrix, which is a common choice in model-based RL, e.g. \citet{vlassis2012bayesian,o2020making}.

Previous works also discuss the limitations of a fixed discount factor and present approaches for more flexible discounting, for example state-dependent \citep{wei2011markov, yoshida2013reinforcement}, state-action-dependent \citep{pitis2019rethinking}, and transition-based discounting \citep{white2017unifying}. We add to this work by demonstrating that discount regularization carries implicit assumptions of equal transition distributions for all state-action pairs and stronger regularization on those with more transition data.

Finally, in \citet{arumugam2018mitigating}, planning is conducted over the set of epsilon-greedy policies rather than deterministic policies. The additional stochasticity during planning prevents tailoring the policy too closely to the model. We show how the work by which we connect discount regularization to a Dirichlet prior by using a weighted average form of the transition matrix applies to epsilon-greedy regularization as well. This connection allows us to directly compare the methods in terms of transition matrix MSE to identify the right method for the environment. Like with Dirichlet prior, it also allows us to compute a state-action-specific parameter to control the amount of regularization.

\section{Background and Notation}\label{sec:background}

\paragraph{Markov Decision Process} We consider a finite, discrete Markov Decision Process (MDP). An MDP $M$ is characterized by $<S,A,R,T,\gamma>$, defined as follows. $S$: State space of size $N$. $A$: Action space. $R(s,a)$: Reward, as a function of state $s$ and action $a$. $T(s,a)$: Transition function, mapping each state-action pair to a probability distribution over successor states. We assume T is unknown and estimated from the data. $\gamma$: Discount factor, $0 \leq \gamma < 1$.

\paragraph{Certainty Equivalence}
Certainty equivalence is a useful approach to offline model-based RL. The agent takes the estimated model as accurate when finding the optimal policy. It separates the estimation of the model from the policy optimization \citep{goodwin1984adaptive}. The maximum likelihood estimate (MLE) is a natural choice for the model estimate, however maximum likelihood solutions can overfit, particularly in the case of small data sets \citep{murphy2012machine}. Often, a better policy is obtained by regularizing the MDP before learning the certainty equivalence policy. 

\section{A Common Form: Regularization as a Weighted Average Transition Matrix}\label{sec:common_form}
The analyses that follow stem from framing each method in a common form: a weighted average transition matrix.  We demonstrate that discount regularization and the posterior mean of the transition matrix under a Dirichlet prior can both be expressed as a weighted average between the MLE transition matrix and a regularization matrix.  

\paragraph{Dirichlet Prior on T} As discussed in Sec.~\ref{sec:relatedworks}, a Dirichlet prior on the transition matrix $T$ functions as a flexible form of regularization. Given a prior on $T$ for state-action pair $(s_n,a_k)$, $T_{\text{prior}}(s_n,a_k) \sim \text{Dirichlet} (\alpha_{n,k,1},...,\alpha_{n,k,N})$, the posterior mean functions as a regularized form. Though simple, this generates several important insights that deepen our understanding and facilitate better regularization.

Let $\langle c_{n,k,1},...,c_{n,k,N} \rangle$ be the transition count data observed from state $s_n$ to states 1 through $N$ under action $a_k$. It can be easily shown that posterior mean of the transition matrix $\hat{T}_{\genfrac{}{}{0pt}{2}{\text{post}}{\text{mean}}}$ is equal to a weighted average of the MLE transition matrix and the mean of the prior:
\begin{align}\label{eqn:dirichlet}
\hat{T}_{\genfrac{}{}{0pt}{2}{\text{post}}{\text{mean}}}(s_n,a_k)=(1-\epsilon) \hat{T}_{MLE}(s_n,a_k) + \epsilon T_{\genfrac{}{}{0pt}{2}{\text{prior}}{\text{mean}}}(s_n,a_k)
\end{align}
where $\epsilon = \frac{\sum_{i=1}^N \alpha_{n,k,i}}{\sum_{i=1}^N c_{n,k,i} + \sum_{i=1}^N \alpha_{n,k,i}}$.\footnote{Please see Appendix \ref{appdx_dirichlet} for derivation.}\footnote{State-action pair index on  $\epsilon_{n,k}$ omitted for readability.}  

\paragraph{Discount Regularization}
Next we show that discount regularization is mathematically equivalent to replacing the transition matrix with the weighted average between that transition matrix and a matrix of zeros. Although this form is unusual as it is not a true transition matrix, we will show that it has utility in relating the amounts of regularization between methods.

To cast discount regularization in certainty-equivalence RL as a weighted average transition matrix, consider the Bellman equation for the value of each state under policy $\pi$, $V^{\pi}= R_{\pi} + \gamma T_{\pi} V^{\pi}$, where the vector $V^{\pi}$ is the value of each state, $R_{\pi}$ is the vector of rewards, and $T_{\pi}$ is the transition matrix, all under policy $\pi$.  Let $\gamma_p < \gamma$ be the planning discount factor, the lower discount factor used for regularization when calculating the certainty-equivalence policy. Then we have the Bellman equation $V^{\pi} = R_{\pi} + \gamma_p T_{\pi} V^{\pi}$. We rewrite the product $\gamma_p T_{\pi}$ from the Bellman equation as the product of true discount factor $\gamma$ and a weighted average matrix: $\gamma_p T_{\pi} = \gamma[(1-\epsilon)T_{\pi} + \epsilon T_{\text{zeros}}]$, where $T_{\text{zeros}}$ is an appropriately sized matrix of zeros and $\epsilon=\frac{\gamma - \gamma_p}{\gamma}$. 

Using this insight, when estimating the transition matrix from data, we can use the following weighted average transition matrix and the true discount factor $\gamma$ for planning in place of the MLE transition matrix and lower discount factor $\gamma_p$. 
\begin{align}\label{eqn:discount_reg}
\hat{T}_{\genfrac{}{}{0pt}{2}{\text{disc}}{\text{reg}}}(s,a) = (1-\epsilon)\hat{T}_{\text{MLE}}(s,a) + \epsilon T_{\text{zeros}}
\end{align}
where $\epsilon = \frac{\gamma -\gamma_p}{\gamma}$.

Eq.~\ref{eqn:discount_reg} also provides another way to view discounting as ``partial termination'' \cite{sutton2018reinforcement}. According to this classic interpretation, the sum of discounted rewards can be viewed as the sum of undiscounted rewards partially terminating with degree 1 minus the discount factor at each step. Similarly, Eq.~\ref{eqn:discount_reg} with $\gamma=1$ represents the agent terminating with probability $\epsilon = 1-\gamma_p$ at each step.

In the following sections, we prove that discount regularization and averaging the transition matrix for each action with a transition matrix in which all rows are the same produce the same optimal policy for the same value of $\epsilon$. Equating our two expressions for $\epsilon$ from Eqs. \ref{eqn:dirichlet} and \ref{eqn:discount_reg} generates a formula for the magnitude of an empirical Bayes prior on $T(s,a)$ implied by any reduced discount factor $\gamma_p$.


\section{Equivalent Policy for Discount Regularization and Dirichlet Prior}\label{sec:unif_disc_equiv}

\subsection{Equivalence Theorem and Proof}

\begin{figure}[ht]
\includegraphics[width=.48\textwidth]{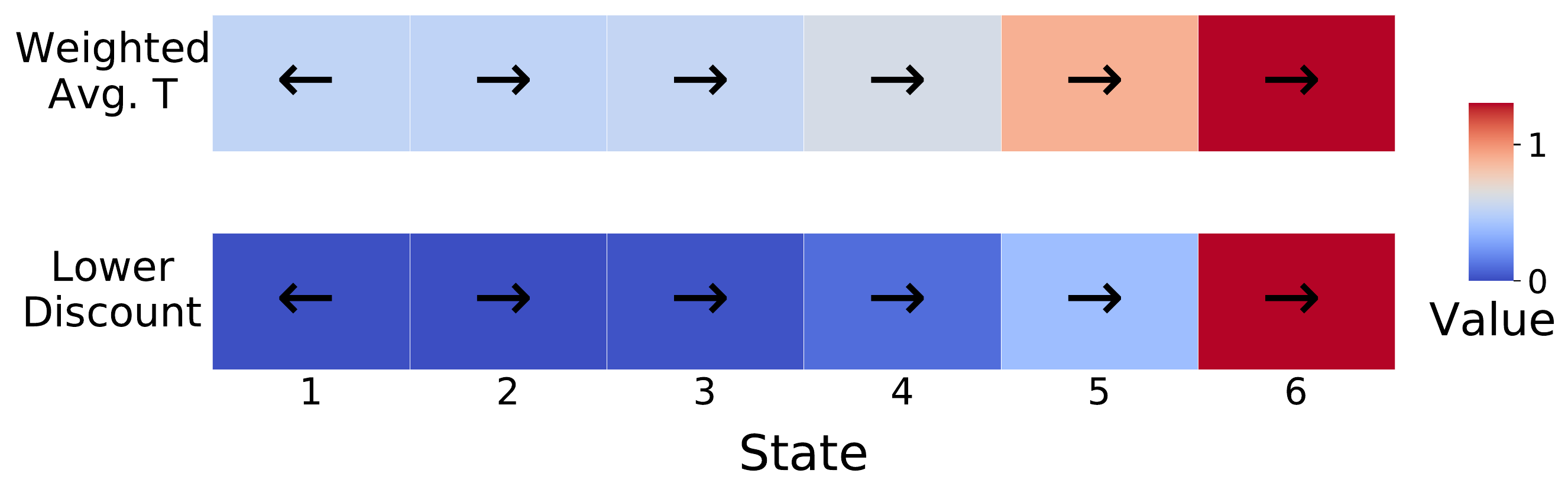}
\centering
\caption{River Swim MDP described in Sec \ref{sec:environments}. Planning with lower discount rate or weighted average $T$ yield different values (colors), but the same optimal policy (arrows).}
\label{fig:heatmap}
\end{figure}

The simplicity of Eqs.~\ref{eqn:dirichlet} and \ref{eqn:discount_reg} allows for direct comparison between the two regularization methods. In fact, discount regularization produces the same optimal policy as averaging the transition matrix with a regularization matrix that is the same for all states and actions when both methods use the same value of $\epsilon$. This result is stated more precisely in Thm.~\ref{the_theorem} and illustrated in Fig.~\ref{fig:heatmap}.

\begin{theorem}\label{the_theorem} 
Let $M_1$ and $M_2$ be finite-state, infinite horizon MDPs with identical state space, action space, and reward function and same discount rate $\gamma < 1$. Let $0 < \epsilon \leq 1$, and let $T_{\text{reg}}(s,a)$ be any matrix used for regularization that is the same for all (s,a) (i.e. identical rows).

If $M_1$ has transition function T and uses discount rate $(1-\epsilon)\gamma$ in planning and $M_2$ has transition function $(1-\epsilon)T+\epsilon T_{reg}$, and uses discount rate $\gamma$ in planning, then $M_1$ and $M_2$ have the same optimal policy.
\end{theorem}

\begin{proof}
The proof is structured as follows. (1) The optimal policy for all MDPs whose Bellman optimality equations differ only by added constant $c$ are the same. (2) The Bellman optimality equation for an MDP in which the transition matrix is regularized by taking its weighted average with a matrix $T_{\text{reg}}$ can be written in terms of a lower discount factor and an added constant. (3) Setting the constant from the previous step to 0, the optimal policy of the resulting MDP is the same as that of the original MDP. (4) The resulting Bellman equation is that of an MDP with the original unregularized transition matrix and reduced discount factor $(1-\epsilon)\gamma$. 

\textbf{(1) } Consider Bellman's optimality equation for any arbitrary state $s$ and action $a$ for an MDP in which constant $c$ is added to every reward $r(s,a)$: $$Q^*(s,a) = r(s,a) + c + \gamma \sum_{s'} T(s,a,s') \text{max}_{a'} Q^*(s',a')$$ It is a known result that the optimal policy of an MDP is not affected by adding the same constant $c$ to all rewards $r(s,a)$. (See, for example, \citet{ng1999policy}: ``constant offets of the reward do not affect the optimal policy when $\gamma < 1$''.)

It follows that the optimal policy  $\pi_{\text{opt}}(s) = \text{argmax}_a Q^*(s,a)$ is the same for all values of $c$. So for all values of constant $c$, the MDP with the Bellman optimality equation above has the same optimal policy.

\textbf{(2) } Let $T_{reg}(s,a)$ be a transition matrix that is the same for all $(s,a)$. We show that Bellman's optimality equation for a transition matrix regularized by taking its weighted average with the $T_{reg}$ can be written in terms of a scaled discount factor and added constant. 
\begin{align*}
Q^*(s,a) &= r(s,a) + \gamma \sum_{s'}[ ( (1 - \epsilon) T(s,a,s') \\
&+ \epsilon T_{reg}(s,a,s')) \text{max}_{a'} Q^*(s',a')]
\end{align*}
\begin{align*}
= r(s,a) &+ \gamma (1 - \epsilon) \sum_{s'} T(s,a,s') \text{max}_{a'} Q^*(s',a') \\
&+ \gamma \epsilon \sum_{s'} T_{reg}(s,a,s') \text{max}_{a'} Q^*(s',a')
\end{align*}
Letting $c(s,a)= \gamma \epsilon \sum_{s'} T_{reg}(s,a,s') \text{max}_{a'} Q^*(s',a')$, Bellman's optimality equation is:
\begin{align*}
&Q^*(s,a) = r(s,a) + c(s,a) \\
&+ \gamma (1 - \epsilon) \sum_{s'} T(s,a,s') \text{max}_{a'} Q^*(s',a')
\end{align*}
By the assumptions of Thm.~\ref{the_theorem}, $T_{reg}(s,a,s')$ is the same for all $(s,a)$ and is therefore a function of $s'$ only. $\text{max}_{a'} Q^*(s',a')$ is also a function of $s'$ only. Therefore $c(s,a)$ is actually a constant number, which we can call $c$.
\begin{align*}
c&=\gamma \epsilon \sum_{s'} \underbrace{T_{reg}(s,a,s')}_{\text{func. of s' only}} \underbrace{\text{max}_{a'}Q^*(s',a')}_{\text{func. of s' only}} = \text{constant}
\end{align*}

\textbf{(3) } By (1), replacing $c$ with 0, the resulting new MDP with Bellman optimality equation
$$Q^*(s,a) = r(s,a)+ \gamma (1 - \epsilon) \sum_{s'} T(s,a,s') \text{max}_{a'} Q^*(s',a')$$
has the same optimal policy.

\textbf{(4) }
This resulting Bellman equation in (3) is that of the MDP with the original, unregularized transition matrix $T(s,a,s')$ and discount factor $\gamma(1-\epsilon)$. Therefore, the MDP with discount rate $\gamma$ and transition matrix $(1-\epsilon)T(s,a,s') + \epsilon T_{reg}(s,a,s')$ and the MDP with discount rate $\gamma (1-\epsilon)$ and transition matrix $T(s,a,s')$ have identical optimal policies.
\end{proof}


Thm.~\ref{the_theorem} provides a deeper understanding of how discount regularization functions. At maximum regularization, $\gamma_p = 0$ or equivalently $\epsilon=1$, it unites two views of the relationship between bandits and MDPs. One common view of a contextual bandit is an MDP with $\gamma$ = 0 \cite{agarwal2019reinforcement}. Alternatively, a contextual bandit is an MDP in which ``the transition probability is identical... for all states and actions'' \citep{zanette2018problem}. Our proof extends this equivalence beyond the bandit setting to all amounts of regularization.  

Thm.~\ref{the_theorem} also reveals the limitations of discount regularization. First, the regularization matrix is the same regardless of the state and action, so it will be biased in environments where transition probabilities vary greatly based on the state and/or the action.  Furthermore, as we demonstrate in the next section, this theorem leads to the result that discount regularization provides stronger regularization on state-action pairs with more data.

\subsection{Dirichlet Prior Implied by Discount Regularization}

We showed that discount regularization produces the same optimal policy as averaging the transition matrix with any matrix that is the same for all states and actions. Recall from Sec.~\ref{sec:common_form} that a Dirichlet prior on $T(s,a)$ also results in a weighted average transition matrix form.
In this section, we further expand on this relationship and will see that using state-action visitation rates from the data allows us to produce an empirical Bayes prior on $T(s,a)$ that results in the same optimal policy as discount regularization.

Using the equivalence in Thm.~ \ref{the_theorem}, we derive the prior magnitude that produces the same optimal policy for any planning discount rate. Since discount regularization employs the same planning discount rate and consequently the same value of $\epsilon$ for every state-action pair, the prior that produces an equivalent policy also has the same value of $\epsilon$ at every state-action pair. Using this equivalence and the two separate formulas for $\epsilon$ in Eqs.~ \ref{eqn:dirichlet} and \ref{eqn:discount_reg} yields a formula for the magnitude of prior implied by any value of planning discount factor $\gamma_p$. Setting the two formulas for $\epsilon$ equal and solving for $\sum_{i=1}^N \alpha_{n,k,i}$, we see that a lower planning discount factor implies a prior whose magnitude depends on the number of transitions from $(s_n,a_k)$ in the data. \footnote{Please see Appendix \ref{appx_dirich} for details.} 
\begin{equation}
\sum_{i=1}^N \alpha_{n,k,i} = \left(\frac{\gamma - \gamma_p}{\gamma_p}\right) \sum_{j=1}^N c_{n,k,j}
    \label{eqn:equivalence}
\end{equation}
In the case of a uniform prior, which we take as the example prior in our simulations, the magnitude simplifies to $$\alpha_{n,k,i} = \left(\frac{\gamma - \gamma_p}{\gamma_p}\right) \frac{\sum_{j=1}^N c_{n,k,j}}{N} \forall i$$

\begin{figure}[ht]
\includegraphics[width=.48\textwidth]{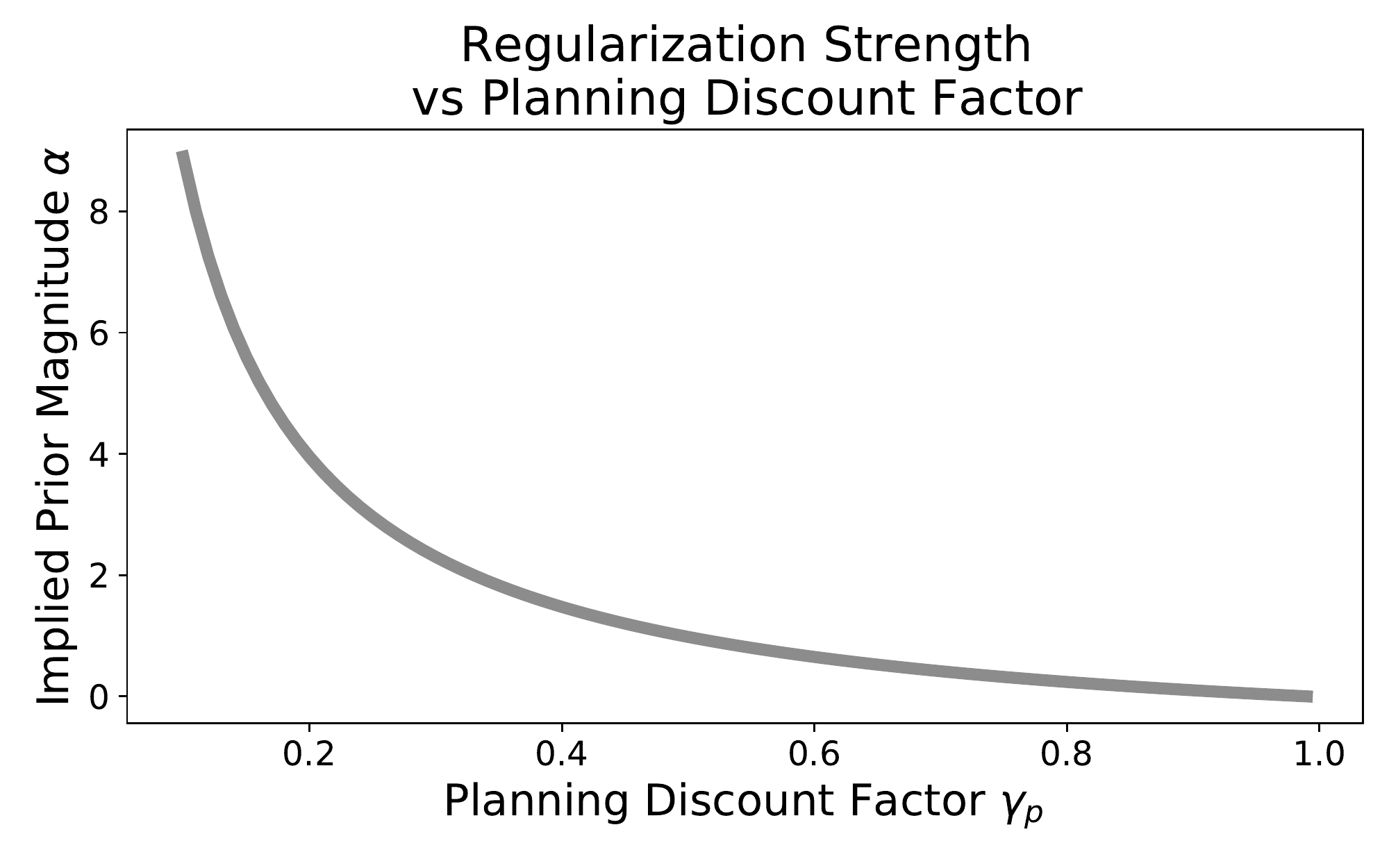}
\centering
\caption{Magnitude of uniform Dirichlet prior implied by planning discount factor $\gamma_p$ for MDP with 10 states, 20 transition observations per state, and $\gamma$ = 0.99.} 
\label{fig:horizon_v_reg}
\end{figure}

The relationship between uniform prior magnitude $\alpha_{n,k,i}$ and planning discount factor $\gamma_p$ for an individual state-action pair is illustrated in Fig.~\ref{fig:horizon_v_reg}. Furthermore, Eq.~\ref{eqn:equivalence} shows us that, for any planning discount factor $\gamma_p$, the magnitude of the corresponding Dirichlet prior is higher for state-action pairs with more data. In other words, those $(s,a)$ with more observations in the data are regularized more. Especially for data sets with uneven distribution of transition data, it may be better to use a more flexible regularization method. In Sec. \ref{sec:sa_specific_eps}, we use our framework to introduce state-action-specific regularization to mitigate this issue. Note also that the special case of $\gamma_p=0$, the contextual bandit setting, presents an exception as the implied priors for all $(s,a)$ are of infinite magnitude. This case is fundamentally different as the future is not just discounted but rather completely ignored.

\section{State-Action-Specific Regularization without Parameter Tuning}\label{sec:sa_specific_eps}

We exposed in Eq.~\ref{eqn:equivalence} that discount regularization functions like a prior on the transition matrix with a potentially undesirable magnitude. To avoid this behavior, we return to the weighted average form introduced in Sec. \ref{sec:common_form} to derive a formula for state-action-specific regularization. Using this form, we calculate the MSE of the estimated transition matrix and identify the value of regularization paramter $\epsilon$ that minimizes the transition matrix MSE. While we recognize that a good transition matrix estimate does not guarantee a good policy, it is a reasonable step towards that goal. 

We derive the closed-form expression of the MSE for the case of a uniform Dirichlet prior. We take $\text{MSE}(\hat{T}(s,a))$ to be the sum of the MSE of the individual elements.  We provide the derivation using the bias-variance decomposition of MSE in Appendix \ref{appdx_MSE_calcs} and the resulting form below. 
Let $\hat{T}_{\text{unif}}$ be the posterior mean of $T$ under a uniform Dirichlet prior. Then,
\begin{equation}
\label{eqn:mse_unif}
\begin{split}
&\text{MSE}[\hat{T}_{\text{unif}}(s_n,a_k)] = \\
& \sum_{i=1}^N \underbrace{(1-\epsilon)^2 \frac{1}{c_{n,k}}T(s_n,a_k,s_i)(1-T(s_n,a_k,s_i))}_{variance} \\
&+ \underbrace{\epsilon\left(\frac{1}{N}-T(s_n,a_k,s_i)\right)^2}_{bias}
\end{split}
\end{equation}
where $c_{n,k}$ is the number of transition observations starting at state $s_n$ under action $a_k$.
Let $\epsilon^*$ to be the value of the regularization parameter $\epsilon$ calculated by minimizing the MSE equation. Then,
\begin{equation}\label{eqn:eps_star_unif}
\epsilon^*(s_n,a_k) = \frac{K(s_n,a_k)}{K(s_n,a_k) + c_{n,k}}
\end{equation}
where $K(s_n,a_k) = \frac{ \sum \limits_{i=1}^N  T(s_n,a_k,s_i)(1-T(s_n,a_k,s_i))}{\sum \limits_{i=1}^N (\frac{1}{N}-T(s_n,a_k,s_i))^2}$.

The first term of Eq.~\ref{eqn:mse_unif} is the contribution of the MLE's variance to the error, in this case the only source of variance. The second term represents the bias introduced by regularization. The strength of regularization $\epsilon$ controls the trade-off between the bias and variance. The variance is driven by the amount of data $c_{n,k}$ both through its role in setting the amount of regularization $\epsilon^*$ and as a factor inversely impacting the variance term. Both bias and variance are impacted by the true transition distribution $T(s,a)$. A deterministic $T(s,a)$ maximizes bias for a given $\epsilon$, but results in $\epsilon^*=0$ (since $T(s,a,s') (1-T(s,a,s'))=0$ for all $s'$). At the other extreme, a $T(s,a)$ with uniform distribution maximizes the variance for a given $\epsilon$ but has no bias, so we default to $\epsilon^*=1$. Intermediate values of $\epsilon^*$ trade off between bias and variance. Of course in practice, the true transition matrix $T$ is generally not known.

A uniform prior on $T(s,a)$ with state-action-specific parameter $\epsilon^*$ improves upon discount regularization by setting the parameters locally for each state-action pair rather than forcing one global regularization parameter. Furthermore, there is no parameter tuning required, simply a plugin estimate for $T$ (e.g. the MLE).  In practice, we may worry that in the low data regimes in which regularization is required, the estimate of $T$ will not be good enough to estimate $\epsilon^*$. Nonetheless, our empirical examples in Sec.~\ref{sec:simulation} demonstrate that our formula for $\epsilon^*$ leads to a reduction in loss over a single global regularization parameter. 

Note that the state-action-specific parameter $\epsilon^*$ combined with regularization matrix $T_{reg}$ does not map directly to a state-action-specific discount factor. The expression for $c(s,a)$ in the proof of Thm~\ref{the_theorem} must be constant for the two methods to produce the same optimal policy and the state-action-specific discount factor breaks this equivalence.

\section{Simulation Results}\label{sec:simulation}

We have demonstrated that planning under a reduced discount factor functions as a prior on the transition matrix with higher magnitude for state-action pairs with more transition observations. We then proposed a better way to regularize by deriving an explicit formula for a uniform prior that minimizes that transition matrix MSE locally for each state-action pair. Next we confirm our results empirically.  

First we demonstrate that the equality in Thm.~\ref{the_theorem} holds. We then compare the performance of (1) discount regularization, (2) a uniform prior on $T$ with equal magnitude for all state-action pairs, and (3) our state-action-specific regularization on three simple tabular examples and a medical cancer simulator. 

\subsection{Tabular Environments}\label{sec:environments}

We demonstrate our results on three common environments from the RL literature.  The first comes from the initial work proposing discount regularization. We choose this environment to demonstrate the limitations of discount regularization even in an environment where it is known to be beneficial. We choose the other two because of their differences in structure, connectivity, and rewards to ensure that our results hold in diverse environments.

\paragraph{10-State Random Chain} The first environment is a distribution over MDPs and we sample one before generating each data set in the examples that follow. \citet{jiang2015dependence} empirically demonstrated the benefits of discount regularization on this randomly generated 10-state, 2-action MDP.  For each state-action pair, 5 successor states are chosen at random to have nonzero transition probability.  These probabilities are drawn independently from Uniform[0,1] and normalized to sum to one. The rewards are sampled independently from Uniform[0,1]. 

\begin{figure}[h]
\includegraphics[width=.48\textwidth]{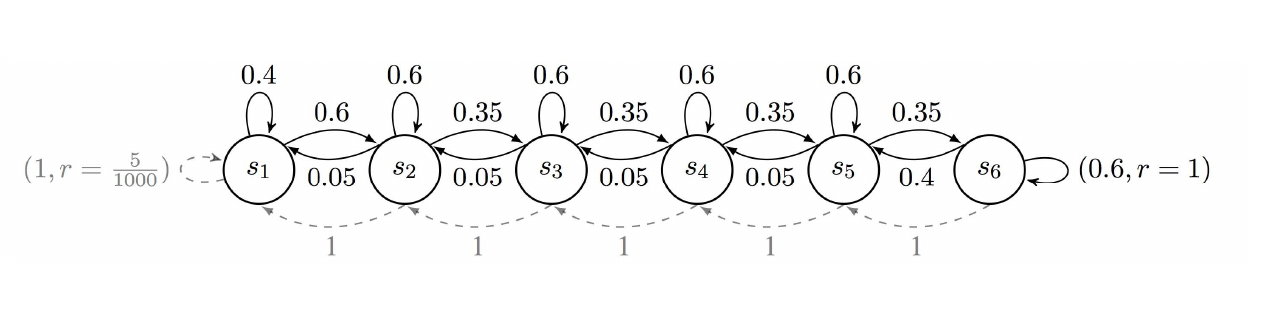}
\centering
\caption{River Swim. Image from \citet{NIPS2013_6a5889bb}.} 
\label{fig:river_swim}
\end{figure}

\paragraph{River Swim}
This common tabular environment described in \citet{NIPS2013_6a5889bb} consists of six states and two actions, as illustrated in Figure \ref{fig:river_swim}. The agent can attempt to swim right ``against the current'' towards the larger reward, or swim left with probability 1 towards the smaller reward.

\begin{figure}[ht]
\includegraphics[width=.33\textwidth]{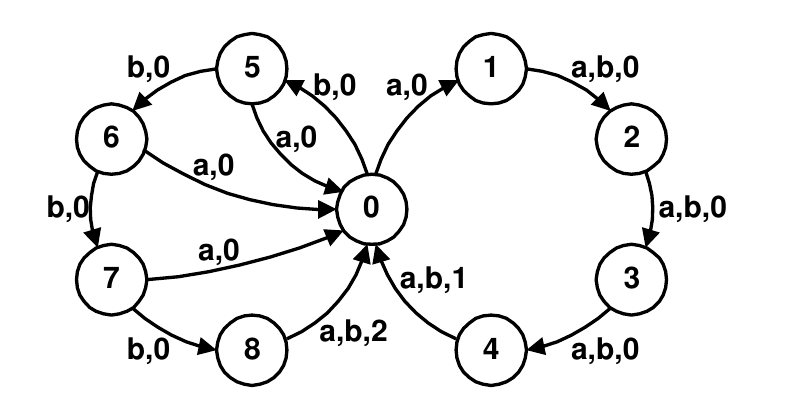}
\centering
\caption{Loop Environment. Image from \citet{strens2000bayesian}.}\label{fig:strens_loop}
\end{figure}

\paragraph{Loop}
The ``Loop'' environment from \citet{strens2000bayesian} consists of nine states forming two loops, joined by a single state. Two actions ``a'' and ``b'' traverse the loops as indicated in Figure \ref{fig:strens_loop}. A reward of 0, 1 or 2 is received at each time step, as indicated in Fig.~\ref{fig:strens_loop}. To add stochasticity to the transitions, we assume that at each time step the agent acts according to the desired action with probability .5 and chooses random between the actions with probability .5.

\subsection{Procedure}
To assess performance in each environment, we follow the procedure in \citet{jiang2015dependence}. We repeatedly sample data sets from the true MDP. (A new MDP is sampled every time in the case of the 10-State Random Chain.) For each, we estimate the transition matrix from the data and assume the reward function is known. Then for a range of regularization strengths ($\epsilon$ or $\gamma$) we regularize the transition matrix separately using (1) discount regularization or (2) a uniform prior with constant magnitude across state-action pairs. We also regularize by (3) a uniform prior with state-action specific parameter. We then calculate the optimal policy.  We compute the loss as the difference between the value of the true optimal policy in the true MDP and the value of the policy found in the estimated, regularized MDP, evaluated in the true MDP. The state-action-specific uniform prior is not dependent on a regularization parameter so we plot the single loss value horizontally.

\subsection{Discount Regularization and Uniform Prior on Transition Matrix Yield Identical Optimal Policies}
\begin{figure}[ht]
\includegraphics[width=.48\textwidth]{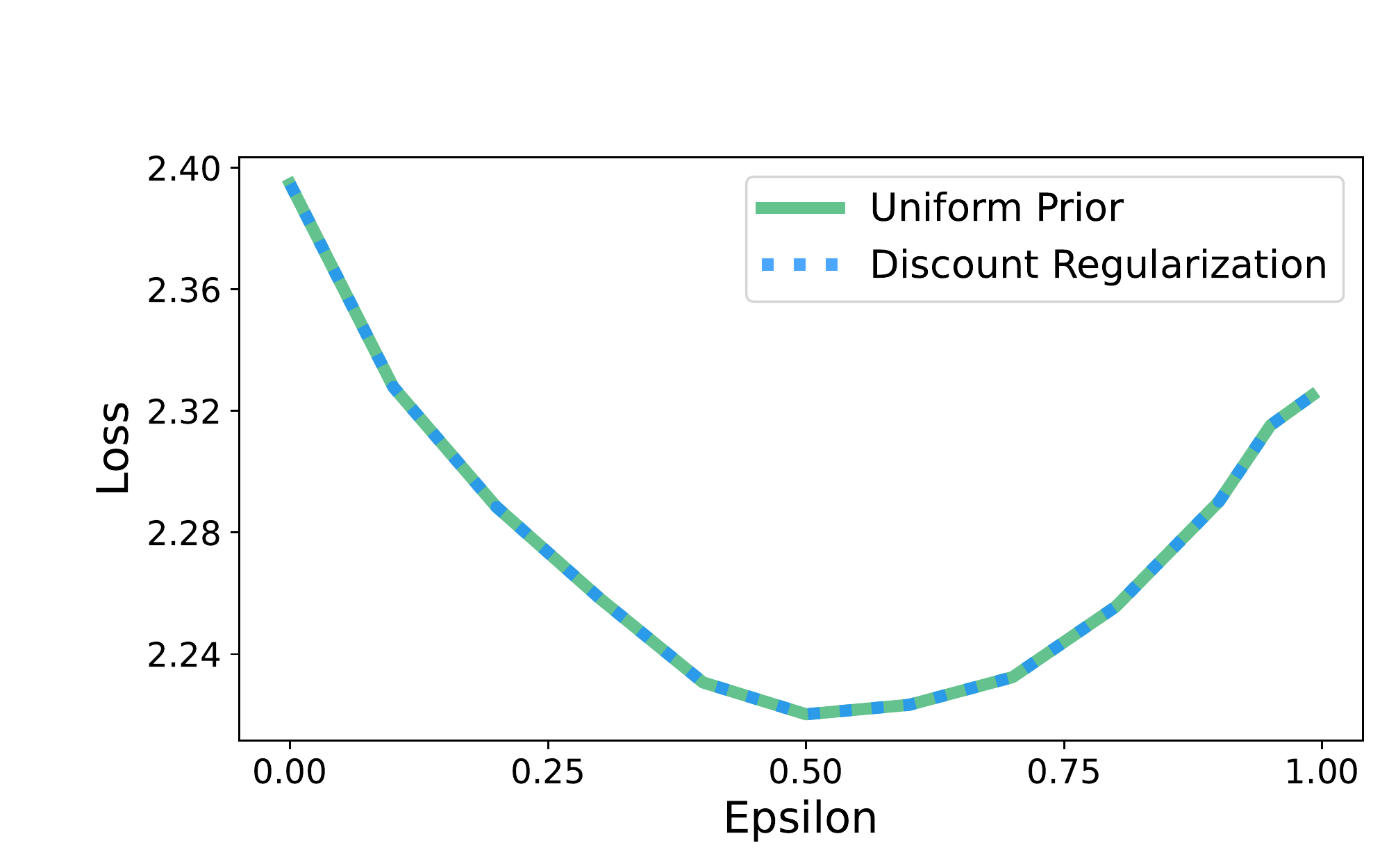}
\centering
\caption{Discount regularization and a uniform prior on the transition matrix result in identical policies when transition count data are equal for all state-action pairs.} 
\label{fig:same_policies}
\end{figure}

First, we empirically confirm our result from Thm. \ref{the_theorem}. When the implied value of $\epsilon$ is the same for all state-action pairs, a uniform prior on $T$ will yield the same optimal policy as a planning discount factor of $\gamma(1-\epsilon)$. As per Eq.~\ref{eqn:equivalence}, we enforce equal $\epsilon$ across state-action pairs by sampling data sets with equal numbers of transition observations across state-action pairs. As demonstrated for the 10-State Random Chain environment in Fig.~\ref{fig:same_policies}, loss is identical for both methods, as is expected for identical policies. 

In the examples that follow, we relax the requirement of equal data across state-action pairs to compare methods under a more realistic data distribution.

\subsection{Exposing Problems with Discount Regularization}

\begin{figure*}
  \begin{subfigure}[b]{0.33\textwidth}
    \includegraphics[width=\textwidth]{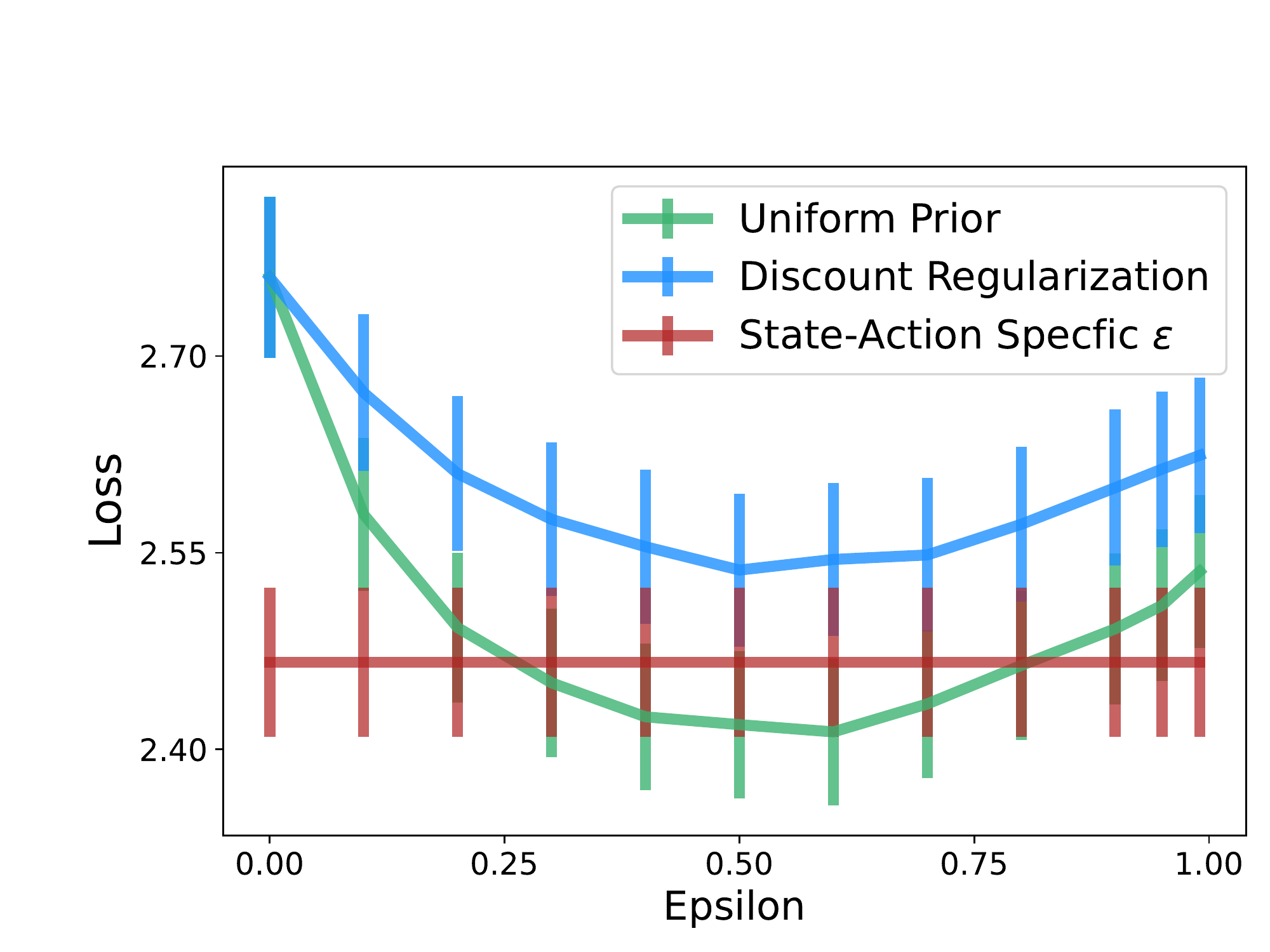}
    \caption{10-State Random Chain}
    \label{fig:1}
  \end{subfigure}
  \begin{subfigure}[b]{0.33\textwidth}
    \includegraphics[width=\textwidth]{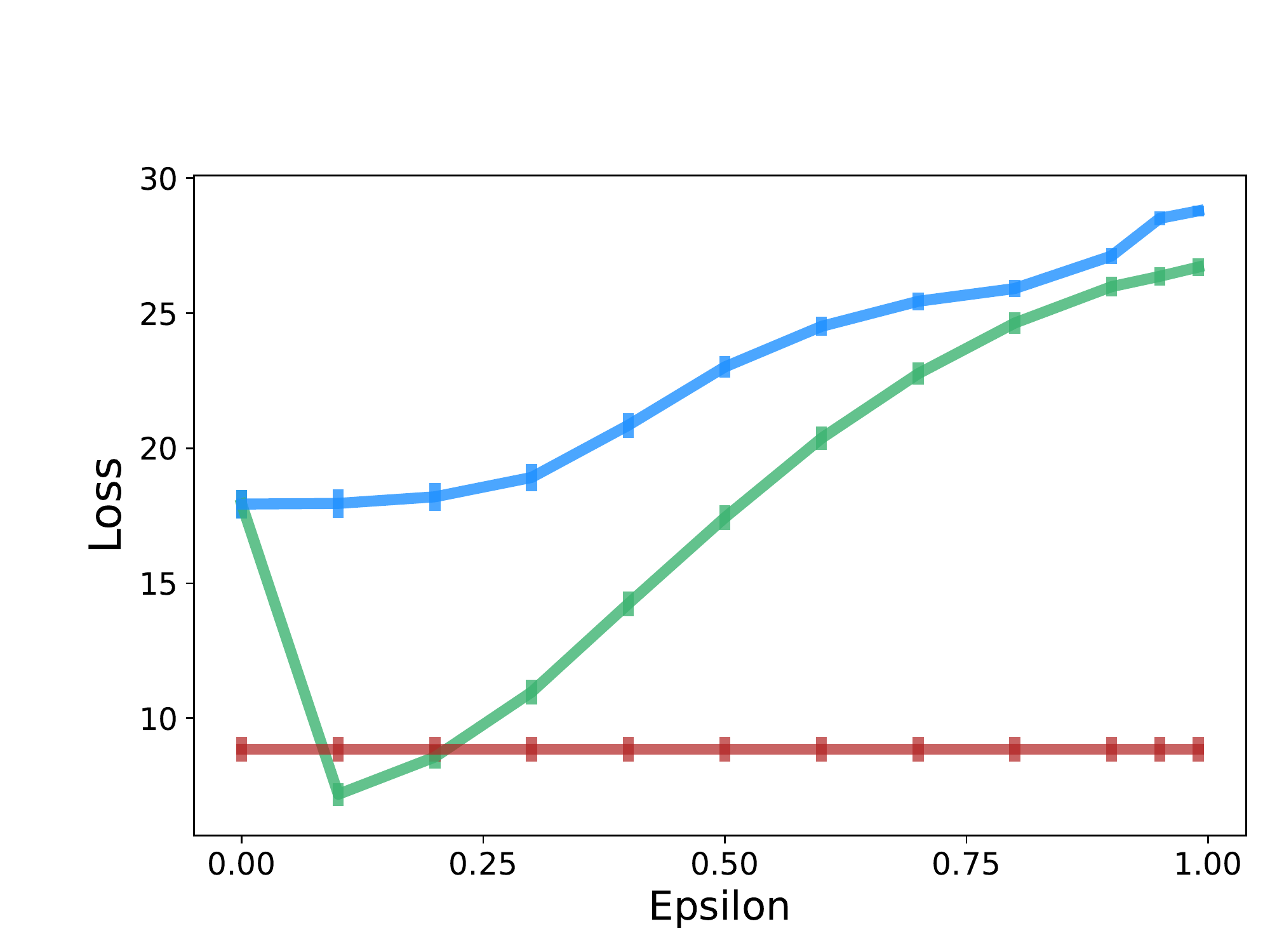}
    \caption{River Swim}
    \label{fig:2}
  \end{subfigure}
  \begin{subfigure}[b]{0.33\textwidth}
    \includegraphics[width=\textwidth]{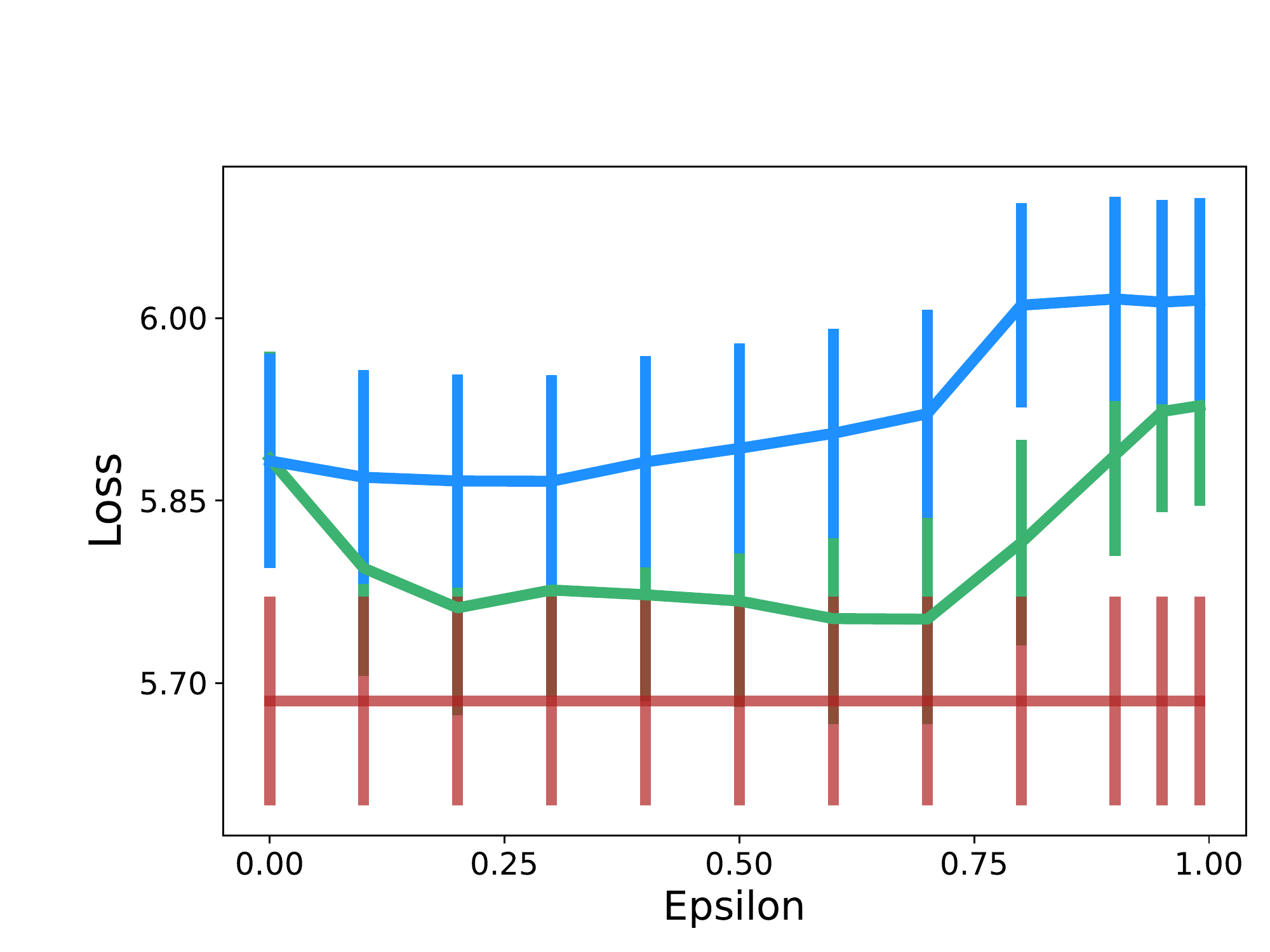}
    \caption{Loop}
    \label{fig:3}
  \end{subfigure}
  \caption{A uniform prior on the transition matrix outperforms discount regularization in all three environments.  A state-action-specific uniform prior performs close to or better than a uniform prior with global regularization parameter $\epsilon$.}
  \label{fig:examples}
\end{figure*}

\paragraph{Discount Regularization performs poorly on data sets with uneven coverage across state-action pairs.}
In real-world conditions, it is unlikely that a data set will have equal numbers of transition observations across state-action pairs.  In this case, recall that discount regularization functions as a prior with higher magnitude for state-action pairs with more data (Eq.~\ref{eqn:equivalence}). We compare this with a uniform prior on the transition matrix with equal magnitude for all state-action pairs.
Fig.~\ref{fig:examples} shows the loss for each method across a range of values of $\epsilon$ (regularization strengths) for the three tabular environments. In these examples, the transition data is generated as tuples $(s,a,r,s')$ with starting state and action chosen uniformly at random, but not enforced to be equal across state-action pairs. Even with transition data that is not heavily skewed away from uniform, the uniform prior with fixed magnitude generates policies that perform better (lower loss) in the true environment across a range of regularization strengths. 

\begin{figure}[ht]
\includegraphics[width=.4\textwidth]{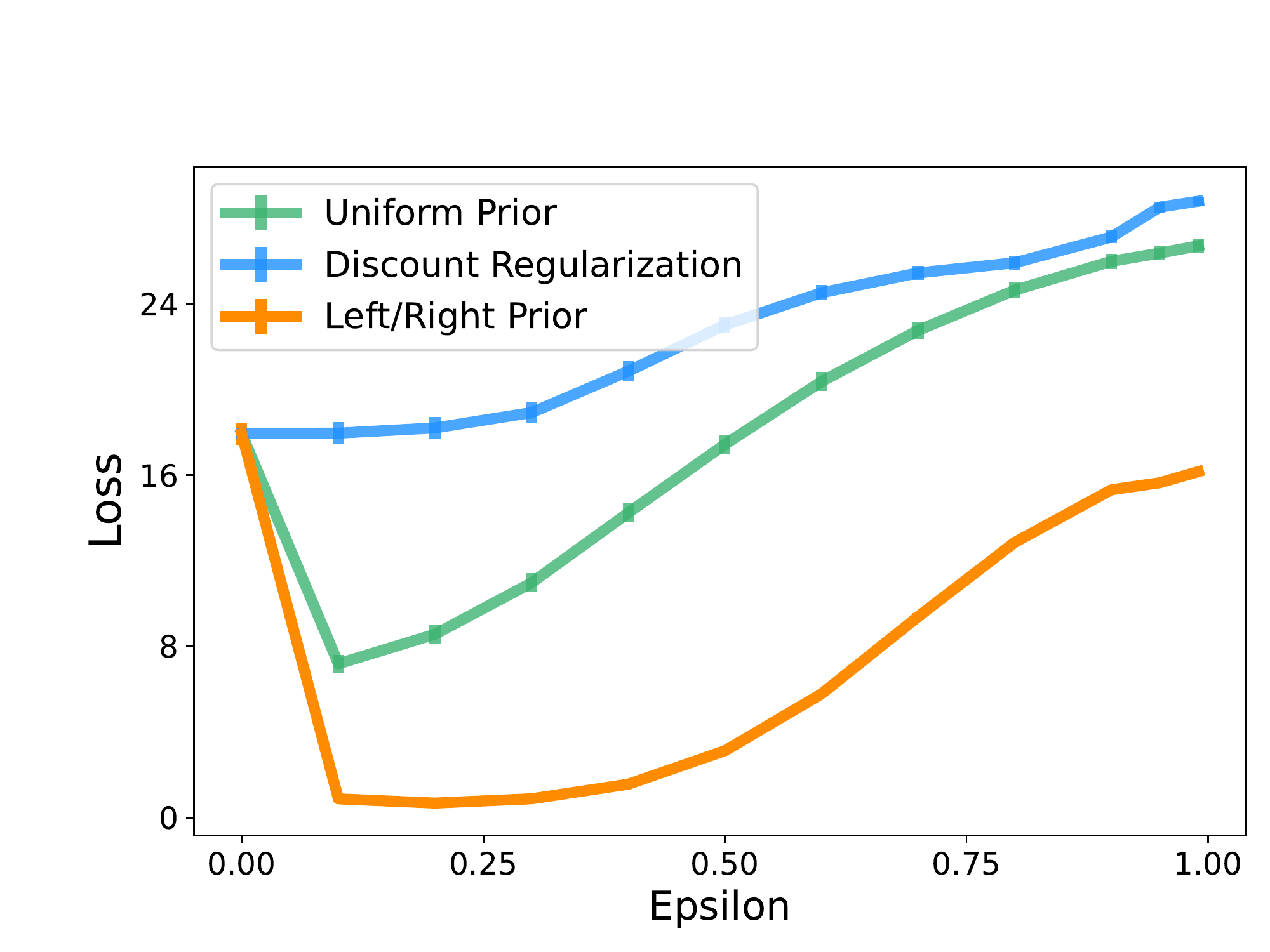}
\centering
\caption{When a uniform prior is not appropriate, a prior
chosen based on expert knowledge of the environment can
perform better.}
\label{fig:non_unif_prior}
\end{figure}

\paragraph{Discount regularization performs poorly when the transition distribution differs greatly across states and/or actions.}
In addition to poor performance in skewed data sets, discount regularization does not perform well in cases where the implied prior, equal for all $(s,a)$, does not match the ground truth. For example, a domain expert may have knowledge that some state transitions are likely or others are impossible. Consider the case of River Swim. If a domain expert knows that Action 1 generally causes the agent to go left and Action 2 generally causes the agent to go right, we may choose a different prior on each action, where the prior on Action 1 determistically moves the agent left and the prior on Action 2 deterministically moves the agent right. Fig.~\ref{fig:non_unif_prior} compares the loss for this deterministic ``left/right prior'' with the other methods. Unsurprisingly, this hand-chosen prior results in lower loss than the methods which assume equal transition distributions for all states and actions.

 \subsection{Simple and Flexible Parameter Tuning}


Performance depends not only on choosing an appropriate regularizer for the data set and environment but also on setting the parameters correctly. We now show how the weighted average transition matrix view of regularization gives a straightforward way to set the regularization level, $\epsilon$, in a state-action-specific way that is easily implemented without cross-validation. 

\paragraph{Our method avoids parameter tuning.} 
Minimizing the transition matrix MSE equation with respect to regularization parameter $\epsilon$ yields an explicit formula for the parameter $\epsilon^*$, Eq.~\ref{eqn:eps_star_unif}. This expression for $\epsilon^*$ depends inversely on the number of transition observations in the data, which allows for reduction in regularization with increased data. The only quantity we lack is an estimate for $T$, which can be approximated by the MLE. Alternatively, we can model $T$ from the data then sample from the posterior, choosing $\epsilon$ to minimize the MSE (Eq.~\ref{eqn:mse_unif}) across the sampled estimates of $T$. This is preferable to cross validation not only because it provides a simple, analytic form, but also because the situations in which regularization is beneficial generally involve few transition observations per state-action pair, resulting in insufficient amounts of data to divide into training and validation sets.

\paragraph{Our method remedies the issue of stronger regularization for state-action pairs with more data.}
Because the formula for $\epsilon^*$ is state-action-specific, it allows the flexibility to adjust the regularization amount separately across state-action pairs with different amounts of data and different transition distributions. This is particularly important as most real-world data sets have uneven distributions and requiring equal regularization across state-action pairs in that case impedes performance. 

Returning to Fig.~\ref{fig:examples}, we demonstrate that our state-action-specific regularization reduces loss without parameter tuning. The horizontal line for ``State-action-specific $\epsilon$'' represents the loss when regularization parameter $\epsilon$ is set separately for each state-action pair. A state-action-specific regularization parameter yields loss that outperforms discount regularization and is close to or outperforms a uniform prior of constant magnitude as well.

In conclusion, we replace cross-validation approaches to parameter tuning with a simple analytical formula that requires only a plug-in estimate of $T$. The regularization parameter is calculated for an individual state-action pair, which allows flexible regularization for uneven data distributions.

\subsection{Cancer Simulation}
We confirm our analysis in a larger, more realistic setting, using a cancer simulator developed by \citet{ribba2012tumor}, as implemented by \citet{gottesman2020interpretable}. The simulator is based on data from patients with a type of tumor called low-grade gliomas (LGG). We use the version for chemotherapy drug TMZ. The structure of the model is based on 21 patients and parameters for the TMZ version are fit using data from 24 patients, with the remaining 96 held out for validation. 

The state space consists of four dimensions: measurements for three different tumor tissue types and the drug concentration. We discretize the states by dividing each dimension into quartiles.  The two actions represent whether or not to administer the chemotherapy drug TMZ at each time step, which represents one month.  The reward at each time step is the reduction in total tumor size from the previous time step, minus a penalty for administering treatment at that time step. In the batch data, treatment at each time step is determined by a draw from the binomial distribution with treatment probability $p$.  We compare regularization methods across a range of parameter choices: amount of stochasticity in the transition between states, magnitude of penalty to the reward for administering chemotherapy, noise in the starting state, and probability $p$ of treatment in the batch data.


As in the previous examples, we compare the loss of the policies generated by discount regularization, a uniform prior on $T$, and state-specific uniform prior.  Across variations in parameters, the two methods with global parameters performed similarly. A risk of both global methods in this case is that if $\epsilon$ is set incorrectly, the loss can be significantly higher. This makes state-action specific regularization particularly compelling, achieving loss near the minimum of all methods with the parameters set globally, but without tuning.

\begin{figure}[ht]
\includegraphics[width=.45\textwidth,height=4.9cm]{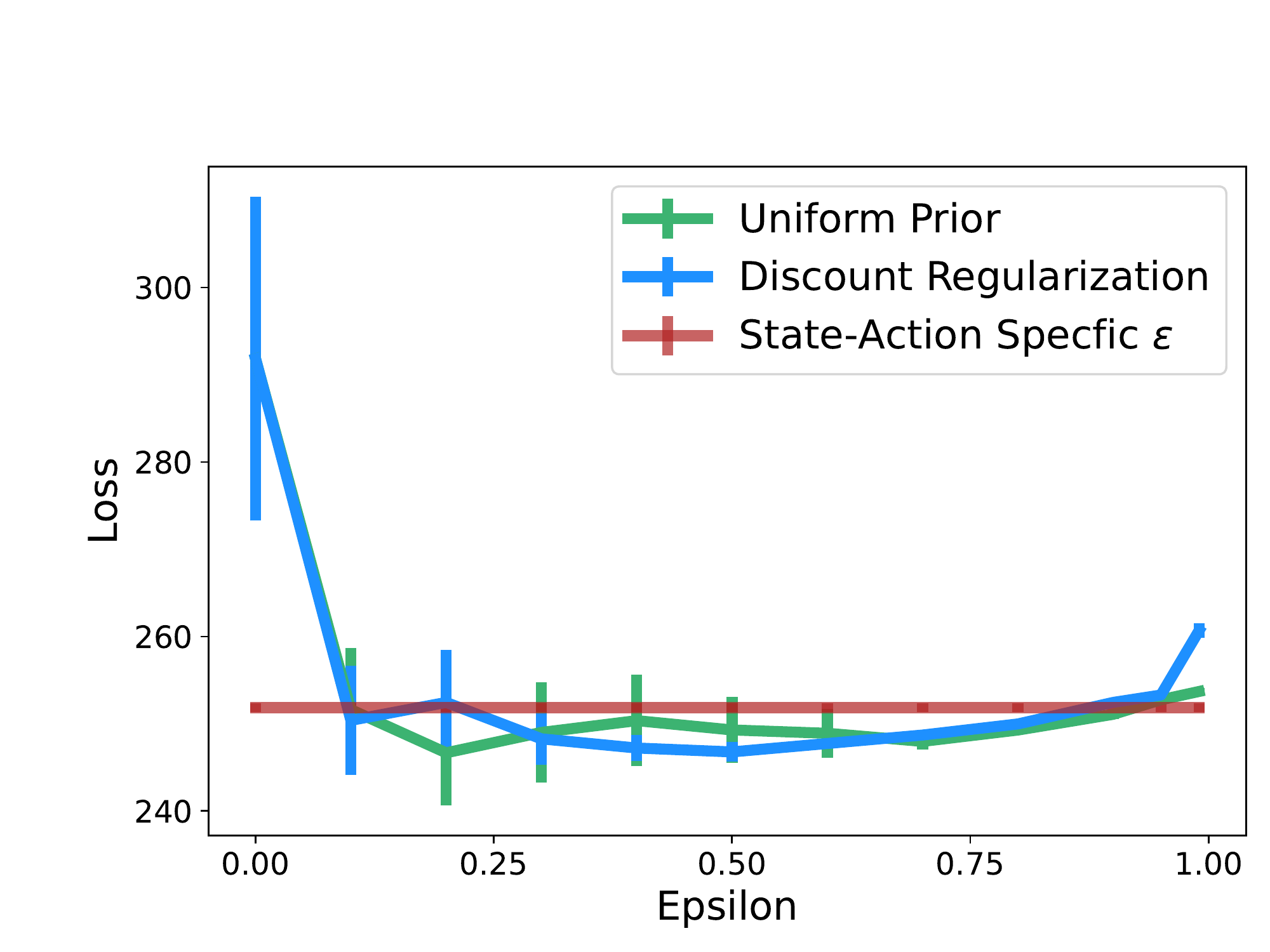}
\centering
\caption{Cancer Simulator: State-action-specific regularization achieves near-minimum loss while avoiding the high loss resulting from incorrectly-set parameters.}\label{fig:cancer}
\end{figure}

\section{Discussion}

\paragraph{Extension to Epsilon-Greedy Regularization} One issue we address is that discount regularization's implicit prior does not match the ground truth of many environments.  As discussed in the example of the ``left/right prior" for River Swim, a uniform prior is not always appropriate. In cases like this, we can extend weighted average form to calculate the MSE and state-action-specific regularization parameter for other methods. One example is epsilon-greedy regularization, described in Sec. \ref{sec:relatedworks}. For this method, the agent treats actions as epsilon-greedy during planning, transitioning according to the transition matrix for the greedy action with probability $(1 -\epsilon)$ otherwise choosing uniformly at random between the transition matrices for all actions. Expressing epsilon-greedy regularization as a weighted average transition matrix allows us to calculate the MSE and state-action-specific parameter, which we do in Appendix \ref{appdx_eg_mse}.  The common form also facilitates comparison between regularization methods.  We can easily compare methods by comparing the regularization matrix that is averaged with the MLE in each case, and choose the regularizer that best matches the environment.

\paragraph{Extension to Model-Free RL}
While we discussed Thm.~\ref{the_theorem} in the context of model-based algorithms, the proof applies to model-free methods as well. To extend our method, we can incorporate a weighted average transition matrix into a model-free method such as Q-learning by drawing random transitions from $T_{reg}$ with probability $\epsilon^*(s,a)$ calculated from Eq.~\ref{eqn:eps_star_unif} and updating the $Q$ function with the random transitions. A sample algorithm and results are presented in Appendix~\ref{appdx_model_free}. The algorithm achieves higher rewards compared to standard Q-learning in many environments.

\paragraph{Limitations} 
As stated in Section~\ref{sec:sa_specific_eps}, our method is based on learning a transition model with low MSE, which does not guarantee a good policy. In other words, it is possible to learn a good policy from a ``bad'' transition model and vice versa, in particular because certain errors in the model may affect the policy more than others. For example, the value equivalence literature demonstrates improved performance by learning a model that minimizes ``value-equivalence loss''-- a loss metric based on the Bellman operators induced by the model-- rather than loss based on maximum likelihood estimation of the model \cite{grimm2020value, grimm2021proper,grimm2022approximate}. We also note that our results are limited to a discrete state space as Thm.~\ref{the_theorem} applies only to the discrete case so we cannot make guarantees of similar results in a continuous state space. Naively discretizing by quantiles poses issues and can be improved upon by methods such as tile coding \cite{stone2001scaling}. Extending our method to the continuous case is a topic of continuing work.


\section{Conclusion}
Discount regularization is a commonly used technique for dealing with noisy and sparse data. Although practitioners believe that they are simply ignoring delayed effects, we revealed through a simple reframing of discount regularization as a weighted average transition matrix that it implicitly assumes a prior on the transition matrix that has the same distribution for all states and actions. Problematically, the magnitude of the prior is higher for state-action pairs with more data. To remedy the issue, we used the weighted average form to derive an explicit formula for the regularization parameter that is calculated locally for each state-action pair rather than globally. Future work will explore the extension of our algorithm to  model-free and continuous state space methodologies. 

\section{Acknowledgements}

Research reported in this work was supported by NIH grants P50DA054039, P41EB028242, and 5R01MH123804-02.

This material is based upon work supported by the National Science Foundation under Grant No. IIS-2007076.  Any opinions, findings, and conclusions or recommendations expressed in this material are those of the author(s) and do not necessarily reflect the views of the National Science Foundation.

\bibliography{main_paper}

\begin{thebibliography}{39}
\providecommand{\natexlab}[1]{#1}
\providecommand{\url}[1]{\texttt{#1}}
\expandafter\ifx\csname urlstyle\endcsname\relax
  \providecommand{\doi}[1]{doi: #1}\else
  \providecommand{\doi}{doi: \begingroup \urlstyle{rm}\Url}\fi

\bibitem[Agarwal et~al.(2019)Agarwal, Jiang, Kakade, and
  Sun]{agarwal2019reinforcement}
Agarwal, A., Jiang, N., Kakade, S.~M., and Sun, W.
\newblock Reinforcement learning: Theory and algorithms.
\newblock \emph{CS Dept., UW Seattle, Seattle, WA, USA, Tech. Rep}, pp.\
  10--4, 2019.

\bibitem[Amit et~al.(2020)Amit, Meir, and Ciosek]{amit2020discount}
Amit, R., Meir, R., and Ciosek, K.
\newblock Discount factor as a regularizer in reinforcement learning.
\newblock In \emph{International conference on machine learning}, pp.\
  269--278. PMLR, 2020.

\bibitem[Arumugam et~al.(2018)Arumugam, Abel, Asadi, Gopalan, Grimm, Lee,
  Lehnert, and Littman]{arumugam2018mitigating}
Arumugam, D., Abel, D., Asadi, K., Gopalan, N., Grimm, C., Lee, J.~K., Lehnert,
  L., and Littman, M.~L.
\newblock Mitigating planner overfitting in model-based reinforcement learning.
\newblock \emph{arXiv preprint arXiv:1812.01129}, 2018.

\bibitem[Asmuth et~al.(2012)Asmuth, Li, Littman, Nouri, and
  Wingate]{asmuth2012bayesian}
Asmuth, J., Li, L., Littman, M.~L., Nouri, A., and Wingate, D.
\newblock A bayesian sampling approach to exploration in reinforcement
  learning.
\newblock \emph{arXiv preprint arXiv:1205.2664}, 2012.

\bibitem[Awasthi et~al.(2022)Awasthi, Guliani, Khan, Vashishtha, Gill, Bhatt,
  Nagori, Gupta, Kumaraguru, and Sethi]{awasthi2022vacsim}
Awasthi, R., Guliani, K.~K., Khan, S.~A., Vashishtha, A., Gill, M.~S., Bhatt,
  A., Nagori, A., Gupta, A., Kumaraguru, P., and Sethi, T.
\newblock Vacsim: Learning effective strategies for covid-19 vaccine
  distribution using reinforcement learning.
\newblock \emph{Intelligence-Based Medicine}, pp.\  100060, 2022.

\bibitem[Cai et~al.(2021)Cai, Grossman, Lin, Sheng, Wei, Williams, and
  Goel]{cai2021bandit}
Cai, W., Grossman, J., Lin, Z.~J., Sheng, H., Wei, J. T.-Z., Williams, J.~J.,
  and Goel, S.
\newblock Bandit algorithms to personalize educational chatbots.
\newblock \emph{Machine Learning}, 110\penalty0 (9):\penalty0 2389--2418, 2021.

\bibitem[Duff(2002)]{duff2002optimal}
Duff, M.~O.
\newblock \emph{Optimal Learning: Computational procedures for Bayes-adaptive
  Markov decision processes}.
\newblock University of Massachusetts Amherst, 2002.

\bibitem[Durand et~al.(2018)Durand, Achilleos, Iacovides, Strati, Mitsis, and
  Pineau]{durand2018contextual}
Durand, A., Achilleos, C., Iacovides, D., Strati, K., Mitsis, G.~D., and
  Pineau, J.
\newblock Contextual bandits for adapting treatment in a mouse model of de novo
  carcinogenesis.
\newblock In \emph{Machine learning for healthcare conference}, pp.\  67--82.
  PMLR, 2018.

\bibitem[Ghavamzadeh et~al.(2015)Ghavamzadeh, Mannor, Pineau, Tamar,
  et~al.]{ghavamzadeh2015bayesian}
Ghavamzadeh, M., Mannor, S., Pineau, J., Tamar, A., et~al.
\newblock Bayesian reinforcement learning: A survey.
\newblock \emph{Foundations and Trends{\textregistered} in Machine Learning},
  8\penalty0 (5-6):\penalty0 359--483, 2015.

\bibitem[Goodwin \& Sin(1984)Goodwin and Sin]{goodwin1984adaptive}
Goodwin, G. and Sin, K.
\newblock Adaptive filtering prediction and control,(book) prentice-hall.
\newblock \emph{Englewood Cliffs}, 6\penalty0 (7):\penalty0 45, 1984.

\bibitem[Gottesman et~al.(2020)Gottesman, Futoma, Liu, Parbhoo, Celi,
  Brunskill, and Doshi-Velez]{gottesman2020interpretable}
Gottesman, O., Futoma, J., Liu, Y., Parbhoo, S., Celi, L., Brunskill, E., and
  Doshi-Velez, F.
\newblock Interpretable off-policy evaluation in reinforcement learning by
  highlighting influential transitions.
\newblock In \emph{International Conference on Machine Learning}, pp.\
  3658--3667. PMLR, 2020.

\bibitem[Grimm et~al.(2020)Grimm, Barreto, Singh, and Silver]{grimm2020value}
Grimm, C., Barreto, A., Singh, S., and Silver, D.
\newblock The value equivalence principle for model-based reinforcement
  learning.
\newblock \emph{Advances in Neural Information Processing Systems},
  33:\penalty0 5541--5552, 2020.

\bibitem[Grimm et~al.(2021)Grimm, Barreto, Farquhar, Silver, and
  Singh]{grimm2021proper}
Grimm, C., Barreto, A., Farquhar, G., Silver, D., and Singh, S.
\newblock Proper value equivalence.
\newblock \emph{Advances in Neural Information Processing Systems},
  34:\penalty0 7773--7786, 2021.

\bibitem[Grimm et~al.(2022)Grimm, Barreto, and Singh]{grimm2022approximate}
Grimm, C., Barreto, A., and Singh, S.
\newblock Approximate value equivalence.
\newblock \emph{Advances in Neural Information Processing Systems},
  35:\penalty0 33029--33040, 2022.

\bibitem[Jiang et~al.(2015)Jiang, Kulesza, Singh, and
  Lewis]{jiang2015dependence}
Jiang, N., Kulesza, A., Singh, S., and Lewis, R.
\newblock The dependence of effective planning horizon on model accuracy.
\newblock In \emph{Proceedings of the 2015 International Conference on
  Autonomous Agents and Multiagent Systems}, pp.\  1181--1189. Citeseer, 2015.

\bibitem[Kolter \& Ng(2009)Kolter and Ng]{kolter2009near}
Kolter, J.~Z. and Ng, A.~Y.
\newblock Near-bayesian exploration in polynomial time.
\newblock In \emph{Proceedings of the 26th annual international conference on
  machine learning}, pp.\  513--520, 2009.

\bibitem[Liao et~al.(2020)Liao, Greenewald, Klasnja, and
  Murphy]{liao2020personalized}
Liao, P., Greenewald, K., Klasnja, P., and Murphy, S.
\newblock Personalized heartsteps: A reinforcement learning algorithm for
  optimizing physical activity.
\newblock \emph{Proceedings of the ACM on Interactive, Mobile, Wearable and
  Ubiquitous Technologies}, 4\penalty0 (1):\penalty0 1--22, 2020.

\bibitem[Murphy(2012)]{murphy2012machine}
Murphy, K.~P.
\newblock \emph{Machine learning: a probabilistic perspective}.
\newblock MIT press, 2012.

\bibitem[Ng et~al.(1999)Ng, Harada, and Russell]{ng1999policy}
Ng, A.~Y., Harada, D., and Russell, S.
\newblock Policy invariance under reward transformations: Theory and
  application to reward shaping.
\newblock In \emph{Icml}, volume~99, pp.\  278--287, 1999.

\bibitem[O'Donoghue et~al.(2020)O'Donoghue, Osband, and Ionescu]{o2020making}
O'Donoghue, B., Osband, I., and Ionescu, C.
\newblock Making sense of reinforcement learning and probabilistic inference.
\newblock \emph{arXiv preprint arXiv:2001.00805}, 2020.

\bibitem[Oh et~al.(2022)Oh, Park, Lee, Kang, and Mo]{oh2022reinforcement}
Oh, S.~H., Park, J., Lee, S.~J., Kang, S., and Mo, J.
\newblock Reinforcement learning-based expanded personalized diabetes treatment
  recommendation using south korean electronic health records.
\newblock \emph{Expert Systems with Applications}, 206:\penalty0 117932, 2022.

\bibitem[Osband et~al.(2013)Osband, Russo, and Van~Roy]{NIPS2013_6a5889bb}
Osband, I., Russo, D., and Van~Roy, B.
\newblock (more) efficient reinforcement learning via posterior sampling.
\newblock In Burges, C., Bottou, L., Welling, M., Ghahramani, Z., and
  Weinberger, K. (eds.), \emph{Advances in Neural Information Processing
  Systems}, volume~26. Curran Associates, Inc., 2013.
\newblock URL
  \url{https://proceedings.neurips.cc/paper/2013/file/6a5889bb0190d0211a991f47bb19a777-Paper.pdf}.

\bibitem[Pitis(2019)]{pitis2019rethinking}
Pitis, S.
\newblock Rethinking the discount factor in reinforcement learning: A decision
  theoretic approach.
\newblock In \emph{Proceedings of the AAAI Conference on Artificial
  Intelligence}, volume~33, pp.\  7949--7956, 2019.

\bibitem[Poggio \& Girosi(1990)Poggio and Girosi]{poggio1990networks}
Poggio, T. and Girosi, F.
\newblock Networks for approximation and learning.
\newblock \emph{Proceedings of the IEEE}, 78\penalty0 (9):\penalty0 1481--1497,
  1990.

\bibitem[Poupart et~al.(2006)Poupart, Vlassis, Hoey, and
  Regan]{poupart2006analytic}
Poupart, P., Vlassis, N., Hoey, J., and Regan, K.
\newblock An analytic solution to discrete bayesian reinforcement learning.
\newblock In \emph{Proceedings of the 23rd international conference on Machine
  learning}, pp.\  697--704, 2006.

\bibitem[Qi et~al.(2018)Qi, Wu, Wang, Tang, and Sun]{qi2018bandit}
Qi, Y., Wu, Q., Wang, H., Tang, J., and Sun, M.
\newblock Bandit learning with implicit feedback.
\newblock \emph{Advances in Neural Information Processing Systems}, 31, 2018.

\bibitem[Ribba et~al.(2012)Ribba, Kaloshi, Peyre, Ricard, Calvez, Tod,
  {\v{C}}ajavec-Bernard, Idbaih, Psimaras, Dainese, et~al.]{ribba2012tumor}
Ribba, B., Kaloshi, G., Peyre, M., Ricard, D., Calvez, V., Tod, M.,
  {\v{C}}ajavec-Bernard, B., Idbaih, A., Psimaras, D., Dainese, L., et~al.
\newblock A tumor growth inhibition model for low-grade glioma treated with
  chemotherapy or radiotherapya tumor growth inhibition model for low-grade
  glioma.
\newblock \emph{Clinical Cancer Research}, 18\penalty0 (18):\penalty0
  5071--5080, 2012.

\bibitem[Ross et~al.(2007)Ross, Chaib-draa, and Pineau]{ross2007bayes}
Ross, S., Chaib-draa, B., and Pineau, J.
\newblock Bayes-adaptive pomdps.
\newblock \emph{Advances in neural information processing systems}, 20, 2007.

\bibitem[Ross et~al.(2011)Ross, Pineau, Chaib-draa, and
  Kreitmann]{ross2011bayesian}
Ross, S., Pineau, J., Chaib-draa, B., and Kreitmann, P.
\newblock A bayesian approach for learning and planning in partially observable
  markov decision processes.
\newblock \emph{Journal of Machine Learning Research}, 12\penalty0 (5), 2011.

\bibitem[Sorg et~al.(2012)Sorg, Singh, and Lewis]{sorg2012variance}
Sorg, J., Singh, S., and Lewis, R.~L.
\newblock Variance-based rewards for approximate bayesian reinforcement
  learning.
\newblock \emph{arXiv preprint arXiv:1203.3518}, 2012.

\bibitem[Stone \& Sutton(2001)Stone and Sutton]{stone2001scaling}
Stone, P. and Sutton, R.~S.
\newblock Scaling reinforcement learning toward robocup soccer.
\newblock In \emph{Icml}, volume~1, pp.\  537--544, 2001.

\bibitem[Strens(2000)]{strens2000bayesian}
Strens, M.
\newblock A bayesian framework for reinforcement learning.
\newblock In \emph{ICML}, volume 2000, pp.\  943--950, 2000.

\bibitem[Sutton \& Barto(2018)Sutton and Barto]{sutton2018reinforcement}
Sutton, R.~S. and Barto, A.~G.
\newblock \emph{Reinforcement learning: An introduction}, pp.\  113.
\newblock MIT press, 2018.

\bibitem[Trella et~al.(2022)Trella, Zhang, Nahum-Shani, Shetty, Doshi-Velez,
  and Murphy]{trella2022reward}
Trella, A.~L., Zhang, K.~W., Nahum-Shani, I., Shetty, V., Doshi-Velez, F., and
  Murphy, S.~A.
\newblock Designing reinforcement learning algorithms for digital
  interventions: pre-implementation guidelines.
\newblock \emph{Algorithms}, 15\penalty0 (8):\penalty0 255, 2022.

\bibitem[Vlassis et~al.(2012)Vlassis, Ghavamzadeh, Mannor, and
  Poupart]{vlassis2012bayesian}
Vlassis, N., Ghavamzadeh, M., Mannor, S., and Poupart, P.
\newblock Bayesian reinforcement learning.
\newblock \emph{Reinforcement learning}, pp.\  359--386, 2012.

\bibitem[Wei \& Guo(2011)Wei and Guo]{wei2011markov}
Wei, Q. and Guo, X.
\newblock Markov decision processes with state-dependent discount factors and
  unbounded rewards/costs.
\newblock \emph{Operations Research Letters}, 39\penalty0 (5):\penalty0
  369--374, 2011.

\bibitem[White(2017)]{white2017unifying}
White, M.
\newblock Unifying task specification in reinforcement learning.
\newblock In \emph{International Conference on Machine Learning}, pp.\
  3742--3750. PMLR, 2017.

\bibitem[Yoshida et~al.(2013)Yoshida, Uchibe, and
  Doya]{yoshida2013reinforcement}
Yoshida, N., Uchibe, E., and Doya, K.
\newblock Reinforcement learning with state-dependent discount factor.
\newblock In \emph{2013 IEEE third joint international conference on
  development and learning and epigenetic robotics (ICDL)}, pp.\  1--6. IEEE,
  2013.

\bibitem[Zanette \& Brunskill(2018)Zanette and Brunskill]{zanette2018problem}
Zanette, A. and Brunskill, E.
\newblock Problem dependent reinforcement learning bounds which can identify
  bandit structure in mdps.
\newblock In \emph{International Conference on Machine Learning}, pp.\
  5747--5755. PMLR, 2018.

\end{thebibliography}
\bibliographystyle{icml2023}

\newpage
\appendix
\onecolumn

\section{Common Form Derivations}

\subsection{Discount Regularization}\label{appdx_discount}

Consider the matrix form of the Bellman equation, using $\gamma_p$, the lower value of the discount factor used during planning for regularization: $V = R + \gamma_p T V$.  By the steps below, we write the product $\gamma_p T$ from the Bellman equation as the product of true discount factor $\gamma$ and a weighted average matrix.
\begin{align*}
\gamma_p T &=[\gamma - (\gamma - \gamma_p)]T \tag{Add and subtract $\gamma$}\\
\gamma_p T &= \gamma(1 - \frac{(\gamma - \gamma_p)}{\gamma})T \tag{Pull out a factor of $\gamma$.}
\end{align*}
Let $T_{\text{zeros}}$ be an appropriately sized matrix of zeros. Adding $\gamma T_{\text{zeros}}$ to the right hand side does not change the equality.
\begin{align*}
\gamma_p T = \gamma[(1-\frac{\gamma-\gamma_p}{\gamma})T+T_{\text{zeros}}]
\end{align*}
Multiply the $T_{\text{zeros}}$ term inside the parentheses by $\frac{\gamma - \gamma_p}{\gamma}$. $T_{\text{zeros}}$ is all zeros so a multiplier does not affect the equality.
\begin{align*}
\gamma_p T = \gamma[(1-\frac{\gamma-\gamma_p}{\gamma})T+(\frac{\gamma-\gamma_p}{\gamma})T_{\text{zeros}}]
\end{align*}
Let $\epsilon=\frac{\gamma-\gamma_p}{\gamma}$.
\begin{align*}
\gamma_p T = \gamma[(1-\epsilon) T_{true} + \epsilon T_{\text{\text{zeros}}}]
\end{align*}

We have replaced the product of the planning discount factor and the true transition matrix with the product of the true discount factor and a weighted average of the transition matrix and a matrix of zeros.  To put this in our framework, consider regularizing the MLE transition matrix for state-action pair $(s,a)$ via discount regularization, using planning discount factor $\gamma_p$. Using the proof in this section, our regularized estimated transition matrix $\hat{T}(s,a)$, is:
\begin{equation*}
\hat{T}(s,a)= (1-\epsilon)\hat{T}_{MLE}(s,a)+\epsilon T_{zeros}
\label{eq:discount}
\end{equation*}
where $\epsilon=\frac{\gamma-\gamma_p}{\gamma}$.

\subsection{Dirichlet Prior Derivation}\label{appdx_dirichlet}

Assume prior $P_{\text{prior}}(T(s_n,a_k)) = \text{Dirichlet} (\langle \alpha_{n,k,1},...,\alpha_{n,k,N} \rangle )$ on transition matrix $T(s_n,a_k)$ and let $\langle c_{n,k,1},...,c_{n,k,N} \rangle$ be the transition count data observed from state $s_n$ to states 1 through $N$ under action $a_k$. The posterior estimate of $T(s_n,a_k)$ follows a Dirichlet distribution with parameter $\langle c_{n,k,1} + \alpha_{n,k,1},...,c_{n,k,N} + \alpha_{n,k,N} \rangle$ and the posterior mean is:
\begin{align*}
T_{\genfrac{}{}{0pt}{2}{\text{post}}{\text{mean}}}&(s_n,a_k) = \langle \frac{c_{n,k,1}+\alpha_{n,k,1}}{\sum\limits_{i=1}^N c_{n,k,i} + \sum\limits_{i=1}^N \alpha_{n,k,i}},...,\frac{c_{n,k,N}+\alpha_{n,k,N}}{\sum\limits_{i=1}^N c_{n,k,i} + \sum\limits_{i=1}^N \alpha_{n,k,i}} \rangle \\
&= \langle \frac{c_{n,k,1}}{\sum\limits_{i=1}^N c_{n,k,i} + \sum\limits_{i=1}^N \alpha_{n,k,i}},...,\frac{c_{n,k,N}}{\sum\limits_{i=1}^N c_{n,k,i} + \sum\limits_{i=1}^N \alpha_{n,k,i}} \rangle + \langle \frac{\alpha_{n,k,1}}{\sum\limits_{i=1}^N c_{n,k,i} + \sum\limits_{i=1}^N \alpha_{n,k,i}},...,\frac{\alpha_{n,k,N}}{\sum\limits_{i=1}^N c_{n,k,i} + \sum\limits_{i=1}^N \alpha_{n,k,i}} \rangle \\
&= \frac{\sum\limits_{i=1}^N c_{n,k,i}}{\sum\limits_{i=1}^N c_{n,k,i}} \langle \frac{c_{n,k,1}}{\sum\limits_{i=1}^N c_{n,k,i} + \sum\limits_{i=1}^N \alpha_{n,k,i}},...,\frac{c_{n,k,N}}{\sum\limits_{i=1}^N c_{n,k,i} + \sum\limits_{i=1}^N \alpha_{n,k,i}} \rangle \\
&+\frac{\sum\limits_{i=1}^N \alpha_{n,k,i}}{\sum\limits_{i=1}^N \alpha_{n,k,i}} \langle \frac{\alpha_{n,k,1}}{\sum\limits_{i=1}^N c_{n,k,i} + \sum\limits_{i=1}^N \alpha_{n,k,i}},...,\frac{\alpha_{n,k,N}}{\sum\limits_{i=1}^N c_{n,k,i} + \sum\limits_{i=1}^N \alpha_{n,k,i}} \rangle \\
&= \frac{\sum\limits_{i=1}^N c_{n,k,i}}{\sum\limits_{i=1}^N c_{n,k,i} + \sum\limits_{i=1}^N \alpha_{n,k,i}} \langle \frac{c_{n,k,1}}{\sum\limits_{i=1}^N c_{n,k,i}},...,\frac{c_{n,k,N}}{\sum\limits_{i=1}^N c_{n,k,i}} \rangle +\frac{\sum\limits_{i=1}^N \alpha_{n,k,i}}{\sum\limits_{i=1}^N c_{n,k,i} + \sum\limits_{i=1}^N \alpha_{n,k,i}} \langle \frac{\alpha_{n,k,1}}{\sum\limits_{i=1}^N \alpha_{n,k,i}},...,\frac{\alpha_{n,k,N}}{\sum\limits_{i=1}^N \alpha_{n,k,i}} \rangle
\end{align*}

Let $\hat{T}_{\text{MLE}}(s_n,a_k)$ be the MLE of $T(s_n,a_k)$. Then, $\hat{T}_{\text{MLE}}(s_n,a_k)= \langle \frac{c_{n,k,1}}{\sum\limits_{i=1}^N c_{n,k,i}},...,\frac{c_{n,k,N}}{\sum
\limits_{i=1}^N c_{n,k,i}} \rangle$. Let $T_{\genfrac{}{}{0pt}{2}{\text{prior}}{\text{mean}}}(s_n,a_k)$ be the transition matrix implied by the prior for state $s_n$ and action $a_k$, i.e. $T_{\genfrac{}{}{0pt}{2}{\text{prior}}{\text{mean}}}(s_n,a_k)= \langle \frac{\alpha_{n,k,1}}{\sum\limits_{i=1}^N \alpha_{n,k,i}},...,\frac{\alpha_{n,k,N}}{\sum\limits_{i=1}^N \alpha_{n,k,i}} \rangle$.

Using $\hat{T}_{\text{MLE}}(s_n,a_k)$ and $T_{\genfrac{}{}{0pt}{2}{\text{prior}}{\text{mean}}}(s_n,a_k)$, we can write $T_{\genfrac{}{}{0pt}{2}{\text{post}}{\text{mean}}}(s_n,a_k)$ as follows.
\begin{align*}
T_{\genfrac{}{}{0pt}{2}{\text{post}}{\text{mean}}}(s_n,a_k) = \frac{\sum\limits_{i=1}^N c_{n,k,i}}{\sum\limits_{i=1}^N c_{n,k,i} + \sum\limits_{i=1}^N \alpha_{n,k,i}} \hat{T}_{\text{MLE}}(s_n,a_k) + \frac{\sum\limits_{i=1}^N \alpha_{n,k,i}}{\sum\limits_{i=1}^N c_{n,k,i} + \sum\limits_{n=1}^N \alpha_{n,k,i} }T_{\genfrac{}{}{0pt}{2}{\text{prior}}{\text{mean}}}(s_n,a_k)
\end{align*}
Let $\epsilon=\frac{\sum\limits_{i=1}^N \alpha_{n,k,i}}{\sum\limits_{i=1}^N c_{n,k,i} + \sum\limits_{i=1}^N \alpha_{n,k,i}}$. Then we have:
\begin{align*}
\label{appdx:eq_1}
T_{\genfrac{}{}{0pt}{2}{\text{post}}{\text{mean}}}(s_n,a_k)=(1-\epsilon) \hat{T}_{\text{MLE}}(s_n,a_k) + \epsilon T_{\genfrac{}{}{0pt}{2}{\text{prior}}{\text{mean}}}(s_n,a_k)
\end{align*}

\subsubsection{Dirichlet Prior Implied by Discount Regularization}\label{appx_dirich}

The equivalence proof above demonstrates that, for a given value of $\epsilon$, averaging the transition matrix with either a matrix of the discrete uniform distribution or with the matrix of zeros yields the same policy. Since discount regularization applies the same $\epsilon$ to every state-action pair, the two methods are only exactly equivalent when the posterior transition matrix under a uniform prior has the same implied value of $\epsilon$ for all state-action pairs, i.e. $\epsilon = \frac{\sum\limits_{i=1}^N \alpha_{n,k,i}}{\sum\limits_{i=1}^N \alpha_{n,k,i} + \sum\limits_{i=1}^N c_{n,k,i}}$ is the same for all state-action pairs, where $c_{n,k,i}$ and $\alpha_{n,k,i}$ are the transition count and prior from state $n$ to state $i$ under action $k$.

We can use the equivalence of the two methods for the same value of $\epsilon$ to solve for the magnitude of prior that is implied by a given discount factor in discount regularization. This is particularly interesting when $\sum\limits_{i=1}^N c_{n,k,i}$ is unequal across state-action pairs $(s_n,a_k)$. In this case, we can use the equivalence of discount regularization and the uniform prior to compute the different priors implied by discount regularization across state-action pairs.

We will refer to $\sum\limits_{i=1}^N c_{n,k,i}$ as $\sum c_i$ and $\sum\limits_{i=1}^N \alpha_{n,k,i}$ as $\sum \alpha_i$ for brevity. Since we know both methods produce the same optimal policy for equal values of $\epsilon$, we set the formulas for $\epsilon$ under each method equal to one another. We then solve for $\alpha_i$ to get the magnitude of prior that is implied by a given planning discount factor $\gamma_p$.

\begin{align*}
\frac{\sum \alpha_i}{\sum \alpha_i + \sum c_i} &= \frac{\gamma - \gamma_p}{\gamma} \\
\gamma \sum \alpha_i &= (\sum \alpha_i + \sum c_i)(\gamma - \gamma_p) \\
\gamma \sum \alpha_i &= \gamma \sum \alpha_i - \gamma_p \sum \alpha_i + \gamma \sum c_i - \gamma_p \sum c_i \\
\gamma_p \sum \alpha_i &= \sum c_i (\gamma - \gamma_p) \\
\sum \alpha_i &= \sum c_i \frac{\gamma - \gamma_p}{\gamma_p}
\end{align*}

For the uniform distribution, all $\alpha_i$ for a given state are the same, so substitute $\sum \alpha_i = N \alpha_i$.
\begin{align*}
    N \alpha_i&= \sum c_i \frac{\gamma - \gamma_p}{\gamma_p}\\
    \alpha_i & = \frac{\sum c_i}{N} \frac{\gamma - \gamma_p}{\gamma_p}
\end{align*}
So discount regularization functions like the Dirichlet prior: 
\begin{equation}
T_{\text{prior}}(s_n,a_k) \sim \text{Dirichlet}(\frac{\sum c_i}{N} \frac{\gamma-\gamma_p}{\gamma_p},...,\frac{\sum c_i}{N}\frac{\gamma-\gamma_p}{\gamma_p})
\label{?}
\end{equation}
where again $\sum c_i$ is the total number of transitions observed in the data starting at state $s_n$ under action $a_k$.

\section{Uniform Prior MSE Calcuation}\label{appdx_MSE_calcs}

Let $\sum\limits_{j=1}^N c_{n,k,j}$ be number of observations for state-action pair $(s_n,a_k)$ in the data. We drop the index and write as $\sum c_j$ below for readability. Let $T(s_n,a_k)$ be the transition probability distribution under action $a_k$ starting at state $s_n$. $N$ is the number of states in the MDP.

\begin{align*}
\text{MSE}(\hat{T}(s_n,a_k)) &= \sum\limits_{i=1}^N \left( \text{Variance}(\hat{T}(s_n,a_k,s_i)) + \text{Bias}^2(\hat{T}(s_n,a_k,s_i)) \right)
\end{align*}

\begin{align*}
    \text{Variance} (\hat{T}_{\text{unif}}(s_n,a_k,s_i)) &= \text{Variance} \left((1-\epsilon)\hat{T}_{\text{MLE}}(s_n,a_k,s_i) + \epsilon \frac{1}{N} \right) \\
    &= \text{Variance} \left((1-\epsilon)\hat{T}_{\text{MLE}}(s_n,a_k,s_i) \right) + \text{Variance} \left(\epsilon \frac{1}{N} \right) \\
    &= (1-\epsilon)^2 \text{Variance}(\hat{T}_{\text{MLE}} (s_n,a_k,s_i)) \\
    &= (1-\epsilon)^2 \frac{1}{\sum c_j} T(s_n,a_k,s_i) (1 - T(s_n,a_k,s_i)) \\
    \text{Bias}(\hat{T}_{\text{unif}}(s_n,a_k,s_i)) &= \mathbb{E}\left[ \hat{T}_{\text{unif}}(s_n,a_k,s_i) \right] - T(s_n,a_k,s_i) \\
    &= \mathbb{E} \left[(1-\epsilon)\hat{T}_{\text{MLE}}(s_n,a_k,s_i) + \epsilon \frac{1}{N} \right] - T(s_n,a_k,s_i) \\
    &= (1-\epsilon) T(s_n,a_k,s_i) + \epsilon \frac{1}{N} - T(s_n,a_k,s_i) \\
    & = \epsilon \left(\frac{1}{N} - T(s_n,a_k,s_i) \right) \\
    \text{MSE} (\hat{T}_{\text{unif}}(s_n,a_k)) &= \sum\limits_{i=1}^N \left( (1-\epsilon)^2 \frac{1}{\sum c_j} T(s_n,a_k,s_i) (1 - T(s_n,a_k,s_i)) + \epsilon^2 \left(\frac{1}{N} - T(s_n,a_k,s_i) \right)^2 \right)
\end{align*}

To solve for the optimal value of $\epsilon$, set the first derivative equal to 0.
\begin{align*}
\frac{\partial \text{MSE}}{\epsilon} = \sum\limits_{i=1}^N -2(1-\epsilon) \frac{1}{\sum c_j} T(s_n,a_k,s_i) (1 - T(s_n,a_k,s_i)) + 2 \epsilon \left(\frac{1}{N} - T(s_n,a_k,s_i) \right)^2
\\
\sum\limits_{i=1}^N -2 \frac{1}{\sum c_j} T(s_n,a_k,s_i) (1 - T(s_n,a_k,s_i)) + 2 \epsilon \frac{1}{\sum c_j} T(s_n,a_k,s_i) (1 - T(s_n,a_k,s_i)) + 2 \epsilon \left(\frac{1}{N} - T(s_n,a_k,s_i) \right)^2 = 0 \\
\sum\limits_{i=1}^N \epsilon \left[\frac{1}{\sum c_j} T(s_n,a_k,s_i) (1 - T(s_n,a_k,s_i)) + \left(\frac{1}{N} - T(s_n,a_k,s_i) \right)^2 \right] = \sum\limits_{i=1}^N \frac{1}{\sum c_j} T(s_n,a_k,s_i) (1 - T(s_n,a_k,s_i)) \\
\end{align*}

\begin{align*}
    \epsilon &= \frac{\sum\limits_{i=1}^N \frac{1}{\sum c_j} T(s_n,a_k,s_i) (1 - T(s_n,a_k,s_i))}{\sum\limits_{i=1}^N \frac{1}{\sum c_j} T(s_n,a_k,s_i) (1 - T(s_n,a_k,s_i)) + \left(\frac{1}{N} - T(s_n,a_k,s_i) \right)^2} \\
&= \frac{\sum\limits_{i=1}^N T(s_n,a_k,s_i) (1 - T(s_n,a_k,s_i))}{\sum\limits_{i=1}^N T(s_n,a_k,s_i) (1 - T(s_n,a_k,s_i)) + \sum c_j \left(\frac{1}{N} - T(s_n,a_k,s_i) \right)^2} \\
&= \frac{\sum\limits_{i=1}^N [T(s_n,a_k,s_i) (1 - T(s_n,a_k,s_i)) ]/ \sum_{i=1}^N [ (\frac{1}{N} - T(s_n,a_k,s_i) )^2 ]}{\sum\limits_{i=1}^N [ T(s_n,a_k,s_i) (1 - T(s_n,a_k,s_i)) ]/\sum\limits_{i=1}^N [(\frac{1}{N} - T(s_n,a_k,s_i) )^2 ]+ \sum c_j} \\
\end{align*}

\section{Empirical Example Details}\label{appdx_exeriment_setup}

\subsection{Regularization Loss by $\epsilon$}

The following pseudocode summarizes the empirical example depicted in Figures \ref{fig:same_policies} through \ref{fig:cancer} to compare the resulting loss for the the optimal policies found using each regularization method, for a range of values of $\epsilon$.

\begin{algorithm}[H]
   \caption{Regularization Loss Pseudocode}
   \label{alg:example1}
\begin{algorithmic}
   \STATE {\bfseries Input:} \text{MDP, list of $\epsilon$ values, regularization method}
   \FOR{$i=1$ {\bfseries to} [number of data sets]}
   \STATE Generate data set of $n$ transition tuples
   \STATE Estimate $\hat{T}_{\text{MLE}}$ from data
    \FOR{$\epsilon$ in list}
    \STATE Regularize transition matrices by amount $\epsilon$
    \STATE Calculate optimal policy $\pi$ of regularized MDP
    \STATE Calculate loss comparing value of $\pi$ vs. value of true optimal policy in true MDP
    \ENDFOR
   \ENDFOR
    \STATE Average loss by $\epsilon$ value across all data sets
\end{algorithmic}
\end{algorithm}

\paragraph{Additional Details} 
Tuples in batch data are collected with uniform probability across state-action pairs. Loss is calculated as the average difference in value across states for the policy found by value iteration in the estimated, regularized MDP as compared to the optimal policy, both evaluated in the true environment. 

\subsection{State-Specific Regularization Loss}
The pseudocode below describes the procedure for the empirical demonstration of state-specific regularization depicted in Figure ~\ref{fig:examples}.

\begin{algorithm}[H]
   \caption{State-Specific Regularization Loss Pseudocode}
   \label{alg:example2}
\begin{algorithmic}
    \STATE {\bfseries Input:} \text{MDP, list of $\epsilon$ values, regularization method}
    \FOR{$i=1$ {\bfseries to} [number of data sets] }
    \STATE Generate data set of $n$ transition tuples
    \STATE Estimate $\hat{T}_{\text{MLE}}$ from data
    \FOR{each state-action pair $(s,a)$}
        \STATE Calculate $\epsilon^*(s,a)$ that minimizes MSE using equations in Sec. \ref{sec:sa_specific_eps}
        \STATE Regularize transition matrix by taking the weighted average of $\hat{T}_{MLE}(s,a)$ and the appropriate regularization matrix, using weight $\epsilon^*(s,a)$
    \ENDFOR
    \STATE Calculate optimal policy $\pi$ of regularized MDP
    \STATE Calculate loss comparing value of $\pi$ vs. value of true optimal policy in true MDP
   \ENDFOR
    \STATE Average loss by $\epsilon$ value across all data sets
\end{algorithmic}
\end{algorithm}

\paragraph{Additional Details} In the step where we calculate $\epsilon^*$ using the Equation \ref{eqn:mse_unif}, two possibilities are to use $\hat{T}_{\text{MLE}}$ as a plug-in estimator for $T$ or model the distribution of $T$ from using the batch data and sample from the posterior. To generate the plot in Figure \ref{fig:examples}, we did the latter.  We calculated the posterior $Dirichlet(\alpha_1,...\alpha_N)$ from the batch data, then sample repeatedly from that distribution, calculating the MSE for each sample across values of $\epsilon$. We then choose for $\epsilon^*$ the value of $\epsilon$ that had the lowest average loss across samples. 

\section{Extension to Epsilon-Greedy Regularization}\label{appdx_eg}

\subsection{Weighted-Average Transition Matrix Form}\label{appdx_eg_common_form}
With the reward expressed as a function of state only (rather than state and action) is straightforward to write epsilon-greedy regularization in the weighted-average form. 

Under epsilon-greedy action selection, the probability of next state $s'$ given state $s_n$ and greedy action $a_k$ is: 

$P_{\genfrac{}{}{0pt}{2}{\text{eps}}{\text{greedy}}}(s'|s_n,a_k) = P(a_k)P(s'|s_n,a_k)+P(\text{random action})P(s'|s_n\text{, random action})$

$P_{\genfrac{}{}{0pt}{2}{\text{eps}}{\text{greedy}}}(s'|s_n,a_k) = (1-\epsilon) P(s'|s_n, a_k) + \epsilon P(s'|s_n\text{, random action})$

The probability for each action in the last term is equal and their probability distributions are independent, so we can rewrite the term as a sum. 

$P_{\genfrac{}{}{0pt}{2}{\text{eps}}{\text{greedy}}}(s'|s_n,a_k) = (1-\epsilon) P(s'|s_n, a_k) + \frac{\epsilon}{|A|} \sum\limits_{a' \in A}  P(s'|s_n, a')$

By definition, $P(s'|s_n,a_k)$ is defined by transition matrix $T$.

$T_{\genfrac{}{}{0pt}{2}{\text{eps}}{\text{greedy}}}(s_n,a_k,s') = (1-\epsilon) T(s_n,a_k,s') + \frac{\epsilon}{|A|} \sum\limits_{a' \in A}  T(s_n,a',s')$

In this setting, we are estimating $\hat{T}$ by the MLE, so we replace $T$ with $\hat{T}_{\text{MLE}}$.  This relationship holds for all next states, so we can extend the form above to the vector of next states, $\hat{T}_{\genfrac{}{}{0pt}{2}{\text{eps}}{\text{greedy}}}(s,a)$.

$\hat{T}_{\genfrac{}{}{0pt}{2}{\text{eps}}{\text{greedy}}}(s_n,a_k) = (1-\epsilon) \hat{T}_{MLE}(s_n,a_k) + \frac{\epsilon}{|A|} \sum\limits_{a' \in A}  \hat{T}_{MLE}(s_n,a')$

\subsection{Epsilon-Greedy Transition Matrix MSE}\label{appdx_eg_mse}
Again, we decompose MSE into bias and variance and calculate each separately.  As in the previous section, let $\sum\limits_{j=1}^N c_{n,k,j}$ be number of observations for state-action pair $(s_n,a_k)$ in the data. Let $T(s_n,a_k)$ be the transition probability distribution under action $a_k$ starting at state $s_n$. $N$ is the number of states in the MDP.

\begin{align*}
    \text{Variance}(\hat{T}_{\genfrac{}{}{0pt}{2}{\text{eps}}{\text{greedy}}}(s_n,a_k,s_i)) &= \text{Variance}\left((1-\epsilon)\hat{T}_{\text{MLE}}(s_n,a_k,s_i) + \epsilon \frac{1}{|A|} \sum\limits_{m =1}^{|A|} \hat{T}(s_n,a_m,s_i) \right) \\
    &= \text{Variance}\left((1-\epsilon + \frac{\epsilon}{|A|})\hat{T}_{\text{MLE}}(s_n,a_k,s_i) + \epsilon \frac{1}{|A|} \sum\limits_{m \neq k} \hat{T}(s_n,a_m,s_i) \right) \\
    &= \left(1-\epsilon + \frac{\epsilon}{|A|}\right)^2 \text{Variance}(\hat{T}_{\text{MLE}}(s_n,a_k,s_i) + \left(\frac{\epsilon}{|A|}\right)^2 \sum\limits_{m \neq k} \text{Variance}(\hat{T}(s_n,a_m,s_i)) \\
    &= \left(1-\epsilon + \frac{\epsilon}{|A|} \right)^2 \frac{1}{\sum\limits_{j=1}^N c_{n,k,j}}T(s_n,a_k,s_i)(1-T(s_n,a_k,s_i)) \\
    &+ \left(\frac{\epsilon}{|A|} \right)^2 \sum\limits_{m \neq k} \frac{1}{\sum\limits_{j=1}^N c_{n,m,j}}T(s_n,a_m,s_i)(1-T(s_n,a_m,s_i)) \\
    \text{Bias}(\hat{T}_{\genfrac{}{}{0pt}{2}{\text{eps}}{\text{greedy}}}(s_n,a_k,s_i)) &= \epsilon \frac{1}{|A|} \sum\limits_{m \neq k} \left(T(s_n,a_m,s_i)-T(s_n,a_k,s_i) \right) \\
\text{MSE}(\hat{T}_{\genfrac{}{}{0pt}{2}{\text{eps}}{\text{greedy}}}(s_n,a_k)) &= \sum\limits_{i=1}^N \left(1-\epsilon + \frac{\epsilon}{|A|} \right)^2 \frac{1}{\sum\limits_{j=1}^N c_{n,k,j}}T(s_n,a_k,s_i)(1-T(s_n,a_k,s_i)) \\
&+ \sum\limits_{i=1}^N \left(\frac{\epsilon}{|A|} \right)^2 \sum\limits_{m \neq k} \frac{1}{\sum\limits_{j=1}^N c_{n,m,j}}T(s_n,a_m,s_i)(1-T(s_n,a_m,s_i)) \\
&+ \sum\limits_{i=1}^N \left(\frac{\epsilon}{|A|} \right)^2 \left(\sum\limits_{m \neq k} T(s_n,a_m,s_i)-T(s_n,a_k,s_i) \right)^2
\end{align*}

To solve for the value of $\epsilon$ that minimizes the MSE, set the first derivative equal to 0:
\begin{align*}
    \frac{\partial \text{MSE}}{\epsilon} = \sum\limits_{i=1}^N  &2 \left(1-\epsilon + \frac{\epsilon}{|A|} \right) \left(\frac{1}{|A|} - 1 \right) \frac{1}{\sum\limits_{j=1}^N c_{n,k,j}} T(s_n,a_k,s_i) (1-T(s_n,a_k,s_i)) \\
    &+ \frac{2 \epsilon}{|A|^2} \sum\limits_{m \neq k } \frac{1}{\sum\limits_{j=1}^N c_{n,m,j}} T(s_n,a_m,s_i)(1-T(s_n,a_m,s_i)) \\
    &+ \frac{2\epsilon}{|A|^2} \sum\limits_{m \neq k} \left(T(s_n,a_m,s_i) - T(s_n,a_k,s_i) \right)^2 \\
    &= \epsilon [ \left(\frac{1}{|A|} - 1 \right)^2 \frac{1}{\sum\limits_{j=1}^N c_{n,k,j}} T(s_n,a_k,s_i)(1-T(s_n,a_k,s_i)) \\ 
    &+ \frac{1}{|A|^2} \sum\limits_{m \neq k} \frac{1}{\sum\limits_{j=1}^N c_{n,m,j}} T(s,a_m,s_i)(1-T(s_n,a_m,s_i))  \\
    &+ \frac{1}{|A|^2} \sum\limits_{m \neq k} (T(s_n,a_m,s_i)-T(s_n,a_k,s_i))^2] \\
    &+ \left(\frac{1}{|A|}-1 \right)\frac{1}{\sum\limits_{j=1}^N c_{n,k,j}}T(s_n,a_k,s_i)(1-T(s_n,a_k,s_i))
\end{align*}

\begin{align*}
    \epsilon = \frac{ \sum\limits_{i=1}^N (1-\frac{1}{|A|})\frac{1}{\sum\limits_{j=1}^N c_{n,k,j}} T(s_n,a_k,s_i)(1-T(s_n,a_k,s_i))}{\splitfrac{ \sum\limits_{i=1}^N (\frac{1}{|A|} - 1)^2 \frac{1}{\sum\limits_{j=1}^N c_{n,k,j}} T(s_n,a_k,s_i)(1-T(s_n,a_k,s_i)) + \frac{1}{|A|^2} \sum\limits_{m \neq k} \frac{1}{\sum \limits_{j=1}^N c_{n,m,k}} T(s_n,a_m,s_i)(1-T(s_n,a_m,s_i))}{+ \frac{1}{|A|^2} \sum\limits_{m \neq k} (T(s_n,a_k,s_i)-T(s_n,a_m,s_i))^2}}
\end{align*}

To demonstrate dependence on the number of data observations, we put this in the form $\frac{K(s,a)}{K'(s,a)+ \sum c_j}$, similar to the above. To do so, assuming equal observations $C$ across all state-action pairs the above is equal to:

\begin{align*}
    \epsilon &= \frac{\sum\limits_{i=1}^N (1-\frac{1}{|A|})T(s_n,a_k,s_i)(1-T(s_n,a_k,s_i))}{\splitfrac{\sum\limits_{i=1}^N (\frac{1}{|A|} - 1)^2 T(s_n,a_k,s_i)(1-T(s_n,a_k,s_i)) + \frac{1}{|A|^2} \sum\limits_{m \neq k} T(s_n,a_m,s_i)(1-T(s_n,a_m,s_i))}{+ C \frac{1}{|A|^2} \sum\limits_{m \neq k} (T(s_n,a_k,s_i)-T(s_n,a_m,s_i))^2}} \\ 
    &= \frac{ \sum\limits_{i=1}^N [(1-\frac{1}{|A|})T(s_n,a_k,s_i)(1-T(s_n,a_k,s_i))]/[ \frac{1}{|A|^2} \sum\limits_{m \neq k} (T(s_n,a_k,s_i)-T(s_n,a_m,s_i))^2]}{ \sum_{i=1}^N |A|^2 \frac{(\frac{1}{|A|} - 1)^2 T(s_n,a_k,s_i)(1-T(s_n,a_k,s_i))}{\sum\limits_{m \neq k} (T(s_n,a_k,s_i)-T(s_n,a_m,s_i))^2} + \frac{\sum\limits_{m \neq k} T(s_n,a_m,s_i)(1-T(s_n,a_m,s_i))}{ \sum\limits_{m \neq k} (T(s_n,a_k,s_i)-T(s_n,a_m,s_i))^2} + C}
\end{align*}

\subsection{Empirical Results for Epsilon-Greedy Regularization}

\subsubsection{Tabular Examples from Section \ref{sec:simulation}}

\begin{figure}[H]
  \begin{subfigure}[b]{0.33\textwidth}
    \includegraphics[width=\textwidth]{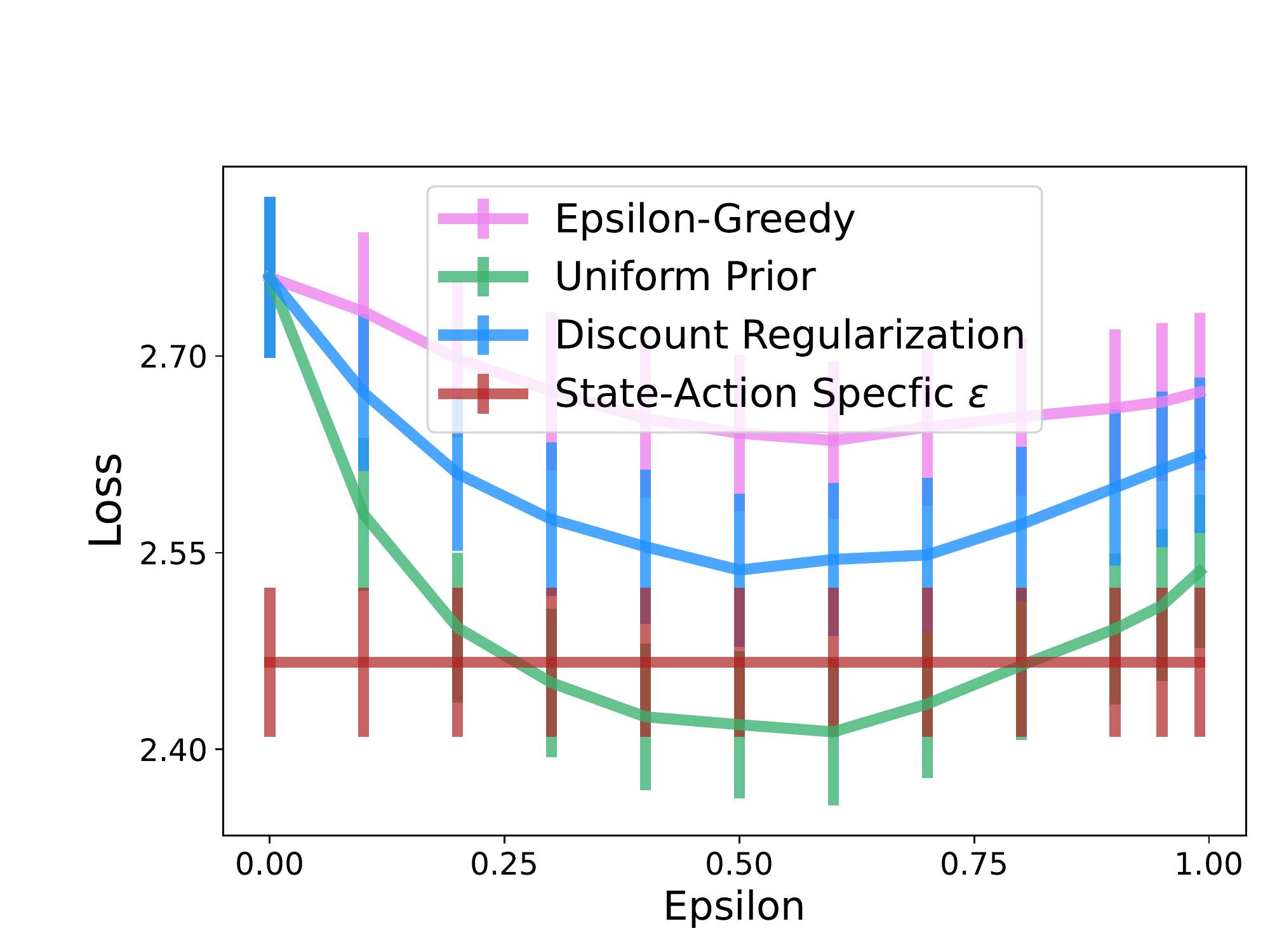}
    \caption{10-State Random Chain}
    \label{fig:1}
  \end{subfigure}
  \begin{subfigure}[b]{0.33\textwidth}
    \includegraphics[width=\textwidth]{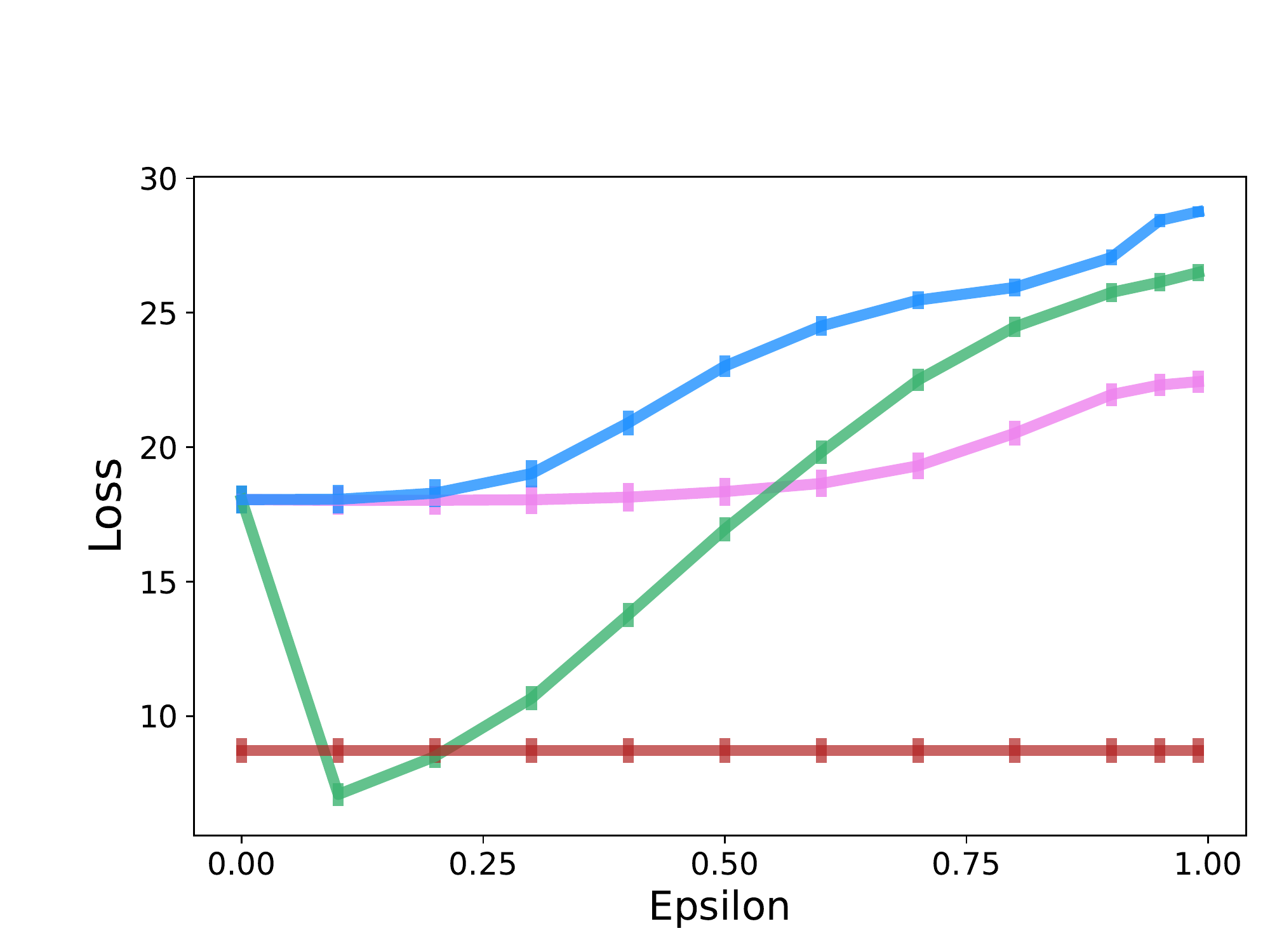}
    \caption{River Swim}
    \label{fig:2}
  \end{subfigure}
  \begin{subfigure}[b]{0.33\textwidth}
    \includegraphics[width=\textwidth]{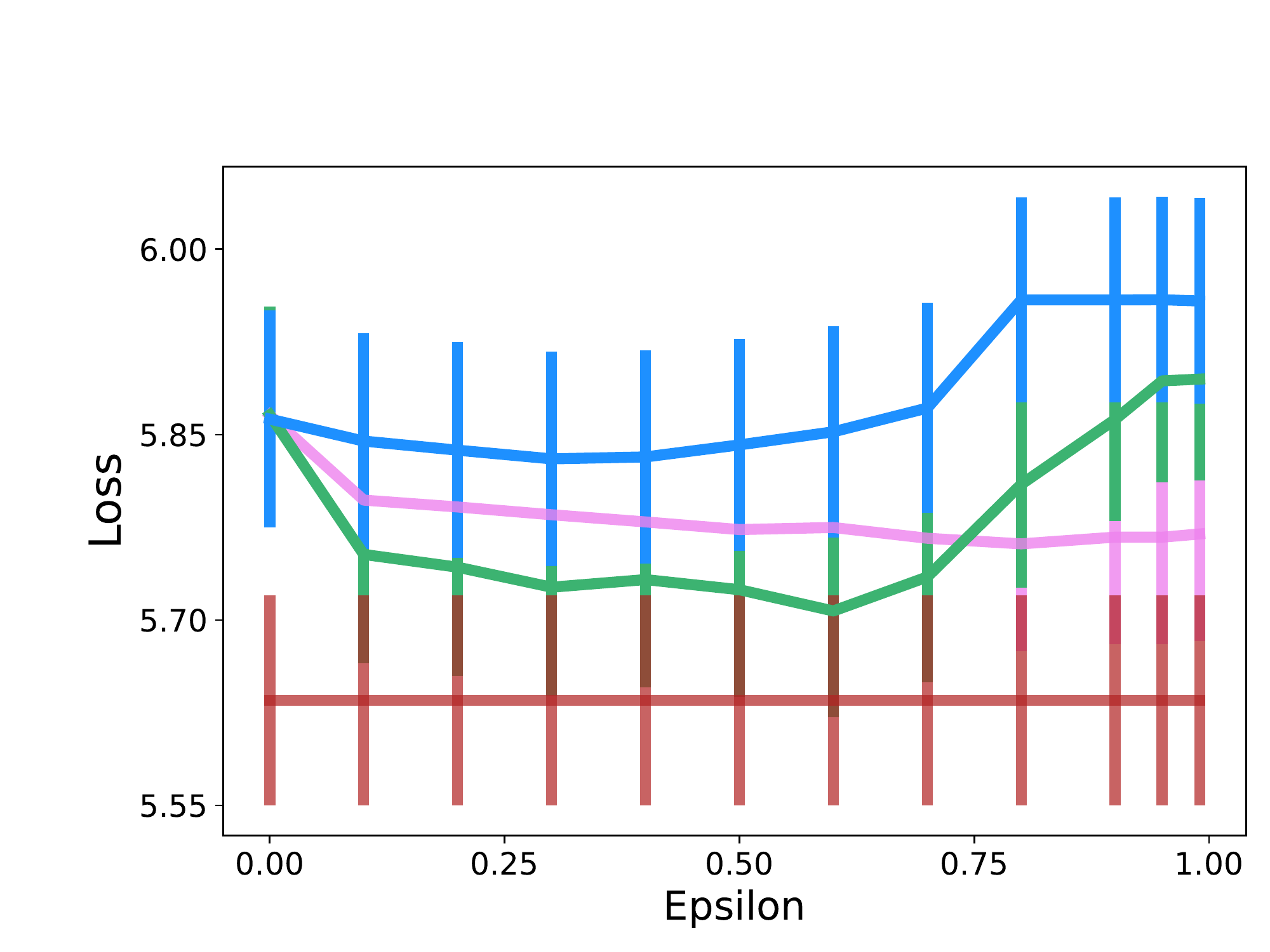}
    \caption{Loop}
    \label{fig:3}
  \end{subfigure}
  \caption{Comparison of epsilon-greedy regularization to uniform prior and discount regularization in our three tabular environments.}
  \label{fig:eg_examples}
\end{figure}

\subsubsection{Controlled Environment}

 \begin{figure}[h]
\includegraphics[width=.75\textwidth]{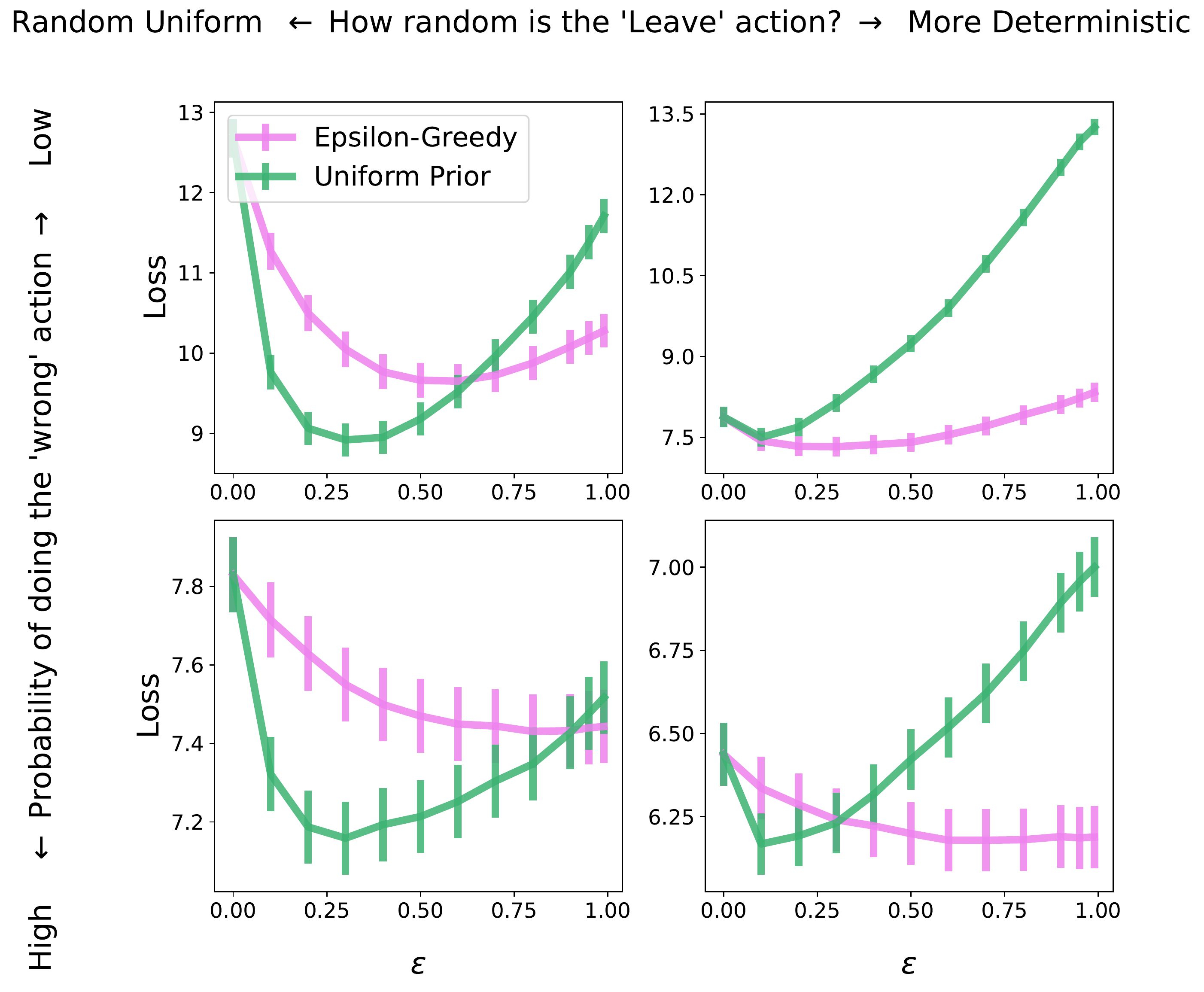}
\centering
\caption{Additional empirical results comparing epsilon-greedy regularization to uniform prior on the environment described in this section. The method whose regularization matrix best aligns with the true environment generates lowest loss.}
\end{figure}

The second set of results come from a loop MDP with 10 states, $\gamma=.99$, and two actions which cause the agent to either stay in or leave the current state, described further below. We define the reward function to be a high reward region in the three adjacent states and zero elsewhere. We vary the transition dynamics along the following two axes to demonstrate the relative performance of different regularizers across environments.

\textbf{Transition Stochasticity, $\kappa$}.   
In our experiments, we utilize a mixture of two transition matrices.  The first is the ``leave'' transiton matrix. We vary the transition dynamics of ``leaving'' the current state, from deterministic (transition from state $s$ to the adjacent state $s+1$ with probability 1) to uniform (transition with uniform probability across states). The second transition matrix is the ``stay'' transition matrix. The ``stay'' transition matrix causes the agent to remain in the current state with probability 0.75, and otherwise transition to a random state.
\begin{align*}
&\text{leave} = \kappa * \text{deterministic }  + (1-\kappa) * \text{uniform} \\
&\text{stay} = 0.75 * \text{identity}  + 0.25 * \text{uniform}
\end{align*}
\textbf{Action Similarity, $\lambda$}. The agent's two actions are generated by mixing the ``stay'' and ``leave'' matrices together in the following way to control the action similarity. 
\begin{align*}
&\text{Action 1 (``probably stay'')} = (1-\lambda)*\text{stay} + \lambda * \text{leave} \\  
&\text{Action 2 (``probably leave'')} = (1-\lambda)*\text{leave} + \lambda * \text{stay} 
\end{align*}
$\lambda$ varies from 0 (distinct actions) to 0.5 (identical actions).

\subsection{Extensions to Model-Free RL}\label{appdx_model_free}

\subsubsection{Sample Algorithm}

\begin{algorithm}[H]
   \caption{Q-learning with State-Action-Specific Regularization}
   \label{alg:model_free}
\begin{algorithmic}
    \STATE Parameters: step size $\alpha \in (0,1]$, regularization matrix $T_{reg}(s,a)$ 
    \STATE Initialize $Q(s,a)=0 \forall (s,a)$
    \FOR{$e=1$ {\bfseries to} [number of episodes] }
    \STATE Choose initial state $s$ randomly.
    \WHILE{step\_counter $<$ [steps per episode]}
        \STATE Choose action $a$ from policy based on $Q(s,a)$ (e.g. epsilon-greedy based on current Q)
        \STATE Calculate $\epsilon^*$ from Eq.~5
        \STATE Draw $x \sim Bernoulli(\epsilon^*)$
        \IF{$x=1$}
            \STATE Draw simulated next step $s_{sim}'$ from $T_{reg}(s,a)$
            \STATE Update Q-function using $s'_{sim}$:
            \STATE $Q(s,a) \leftarrow Q(s,a) + \alpha[r(s,a) + \gamma max_a Q(s'_{sim},A)-Q(s,a)]$
        \ELSIF{$x=0$}    
            \STATE Agent takes action $a$, observes next state $s'$
            \STATE $Q(s,a) \leftarrow Q(s,a) + \alpha[r(s,a) + \gamma max_a Q(s',A)-Q(s,a)]$
            \STATE step\_counter += 1
            \STATE $s \leftarrow s'$
        \ENDIF
    \ENDWHILE
    \ENDFOR
\end{algorithmic}\label{model_free_algo}
\end{algorithm}

\subsubsection{Results}

\begin{figure}[H]
  \begin{subfigure}[b]{0.33\textwidth}
    \includegraphics[width=\textwidth]{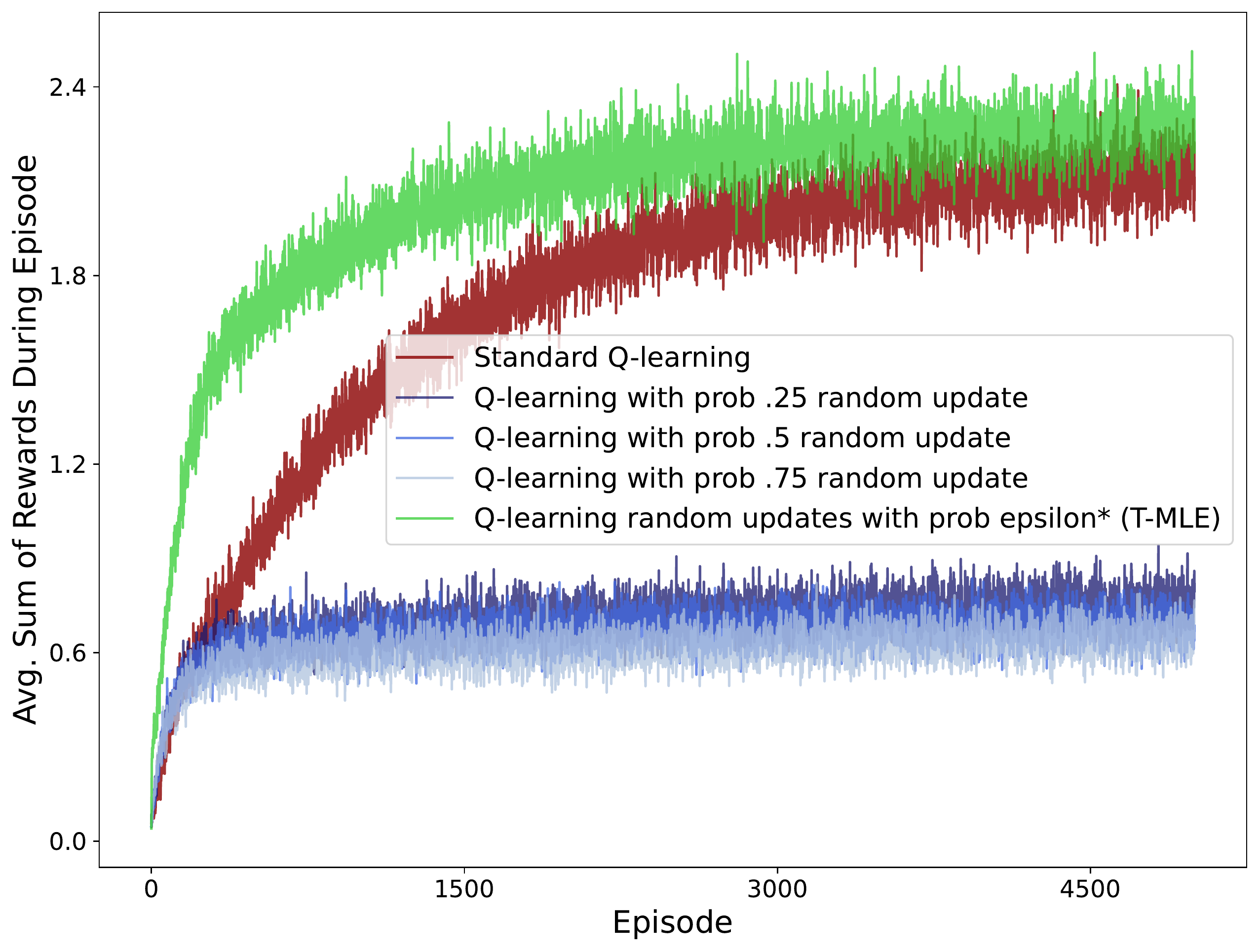}
    \caption{River Swim}
    \label{fig:1}
  \end{subfigure}
  \begin{subfigure}[b]{0.33\textwidth}
    \includegraphics[width=\textwidth]{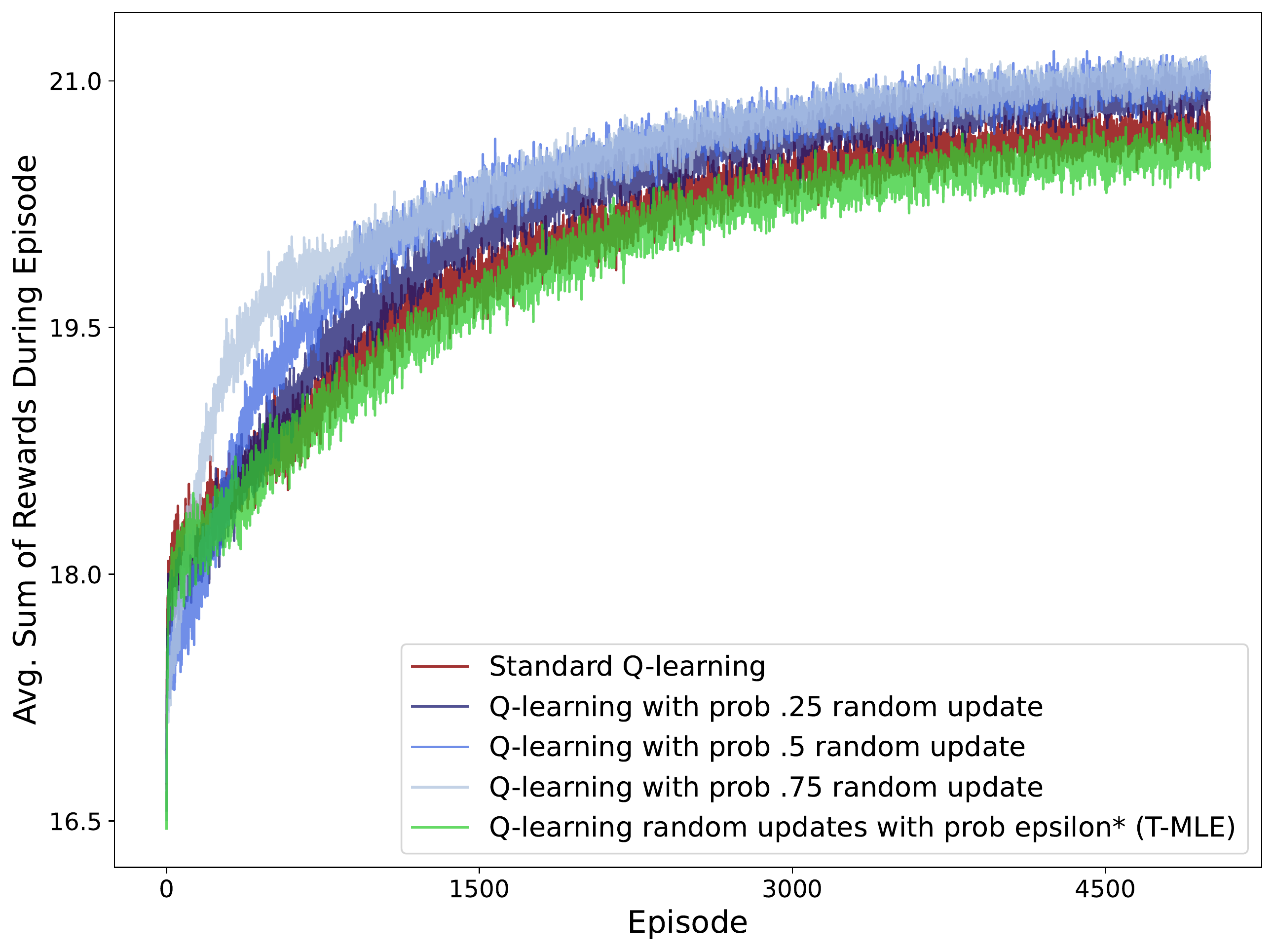}
    \caption{10-State Random Chain}
    \label{fig:2}
  \end{subfigure}
  \begin{subfigure}[b]{0.33\textwidth}
    \includegraphics[width=\textwidth]{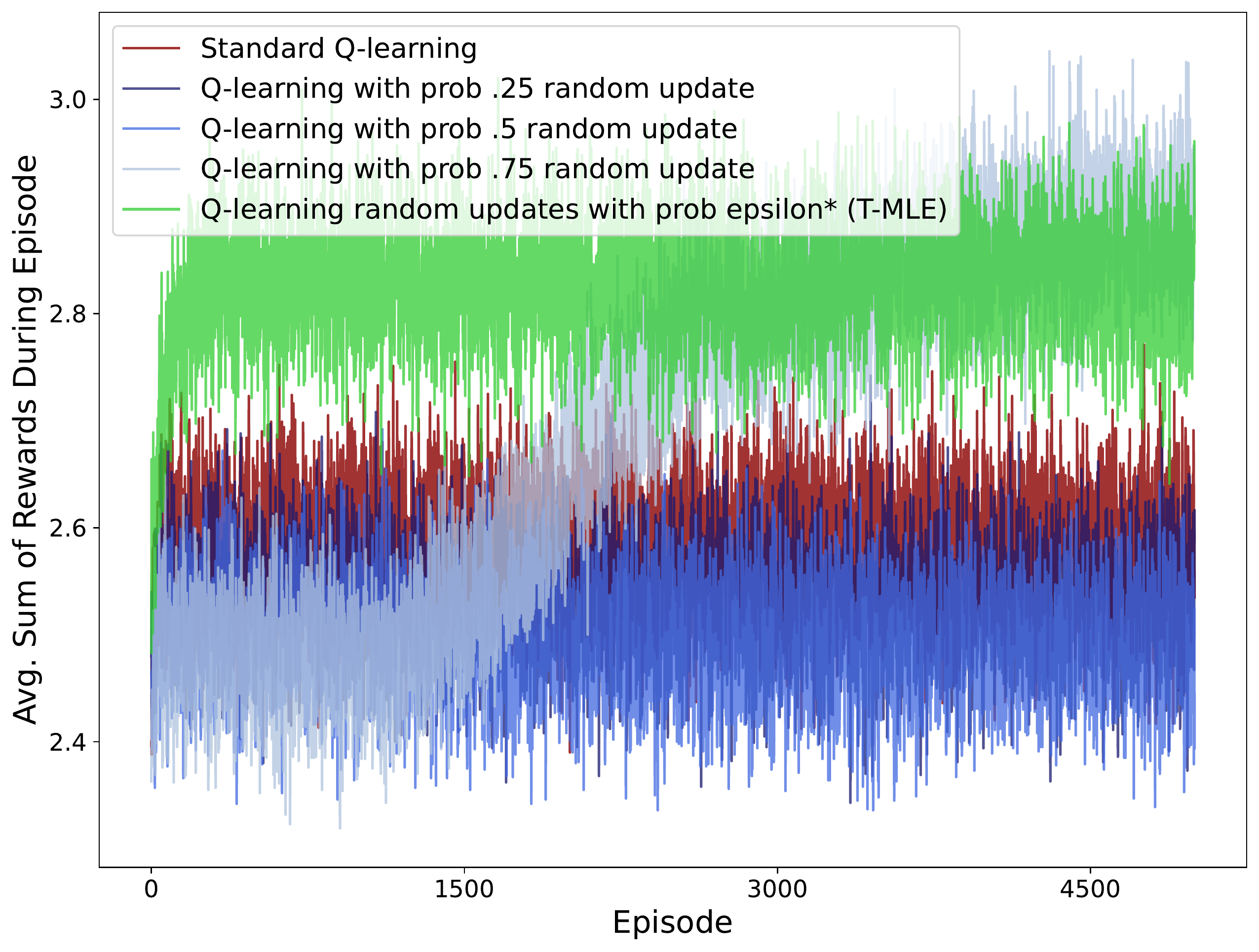}
    \caption{Loop}
    \label{fig:3}
  \end{subfigure}
  \caption{Comparison of epsilon-greedy regularization to standard Q-learning in our three tabular environments. We also include as a baseline our algorithm with $\epsilon^*$ replaced by a constant probability.}
  \label{fig:eg_examples}
\end{figure}

Our simple modified Q-learning algorithm outperforms standard Q-learning on River Swim and Loop environments, but not on the random chain environment.  Understanding where it performs best, why, and improving the extensions to model-free algorithms is a topic of ongoing research.


\end{document}